\documentclass[10pt,twocolumn,letterpaper]{article}
\usepackage[pagenumbers]{westlakeagi}
\usepackage{tcolorbox}
\usepackage{multirow}
\usepackage{array}
\usepackage{colortbl}
\usepackage{algorithm}
\usepackage{algorithmic}
\usepackage[hang,flushmargin]{footmisc}
\usepackage{fontawesome5}

\usepackage[pagebackref,breaklinks,colorlinks,allcolors=wagiblue]{hyperref}
\AtEndPreamble{
  \crefname{figure}{Fig.}{Figs.}
  \crefname{appendix}{App.}{Apps.}
}

\newcommand{\bag}{\texttt{BAG}}
\definecolor{commentcolor}{RGB}{115,115,115}
\definecolor{paramcolor}{rgb}{0.9, 0.2, 0.5}
\definecolor{lightCyan}{rgb}{1,0.95,0.98}
\renewcommand{\eg}{\emph{e.g.},}

\renewcommand{\paragraph}[1]{\vspace{0.7em}\noindent\textbf{#1}}
\newcolumntype{L}[1]{>{\raggedright\arraybackslash}p{#1}}
\newcommand{\dashedmidrule}[1]{%
  \noalign{\vskip1.6pt}%
  \multispan{#1}\leaders\hbox to 4pt{\hss\vrule height.35pt width2pt\hss}\hfill\\%
  \noalign{\vskip1.6pt}%
}

\newcommand{\papertitle}{BAG: Budget-Aware Gating for Diffusion Caching}
\title{\papertitle}

\begin{document}

\twocolumn[{%
  \vspace*{-0.42in}%
  \wagibanner
  \vskip 0.16in
  \begin{tcolorbox}[colback=wagiblue!4, colframe=wagiblue!40,
                    boxrule=0.6pt, arc=10pt,
                    left=22pt, right=22pt, top=14pt, bottom=10pt]
    \begin{center}
      {\LARGE\bf \papertitle\par}
      \vskip 21pt
      {\large
       Tong Zhao$^{1,2}$, \quad
       Mingkun Lei$^{2}$, \quad
       Yucheng Han$^{3}$, \quad
       Chi Zhang$^{2,\dagger}$\par}
      \vskip 6pt
      {$^{1}$Zhejiang University \quad
       $^{2}$AGI Lab, Westlake University \quad
       $^{3}$StepFun\par}
      \vskip 8pt
      {\faGithub\hspace{0.5em}\href{https://github.com/Westlake-AGI-Lab/BAG}{\texttt{https://github.com/Westlake-AGI-Lab/BAG}}\par}
    \end{center}
    \vskip 13pt
    \centerline{\large\bf Abstract}
    \vskip 7pt
    {\it\noindent
Diffusion caching is a lightweight strategy that accelerates Diffusion Transformers (DiTs) by reusing intermediate features across denoising steps, but existing paradigms face a fundamental trade-off: online heuristics lack global budget awareness, whereas static schedules lack instance adaptivity and fail to flexibly adapt to varying runtime budget constraints. To bridge this gap, we present BAG (Budget-Aware Gating), a novel caching policy that unifies global budget pacing with dynamic, instance-adaptive feature reuse. Rather than relying on hand-crafted rules, BAG employs a lightweight gating network that dynamically decides whether to execute a full computation or reuse cached features at each step by jointly conditioning on the budget state and local trajectory feedback. We train this policy via offline-to-online schedule distillation, transferring the decision-making of offline-searched schedules into a compact online gate. Extensive experiments on FLUX.1-dev, Wan2.1, and Qwen-Image-2512 demonstrate that BAG consistently outperforms state-of-the-art caching methods across various speedup tiers while remaining robust across different resolutions, seeds, and guidance scales. Code will be released.\par}
    \vskip 7pt
    {\color{wagiblue!40}\rule{0.3\textwidth}{0.4pt}}\\[2.5pt]
    {\footnotesize $^{\dagger}$Corresponding author.\quad
     Emails: \texttt{\{zhaotong68, chizhang\}@westlake.edu.cn}\par}
  \end{tcolorbox}
  \vskip 0.35in
}]

\section{Introduction}
\label{sec:intro}
\begin{figure*}[t]
\centering
\includegraphics[width=\textwidth]{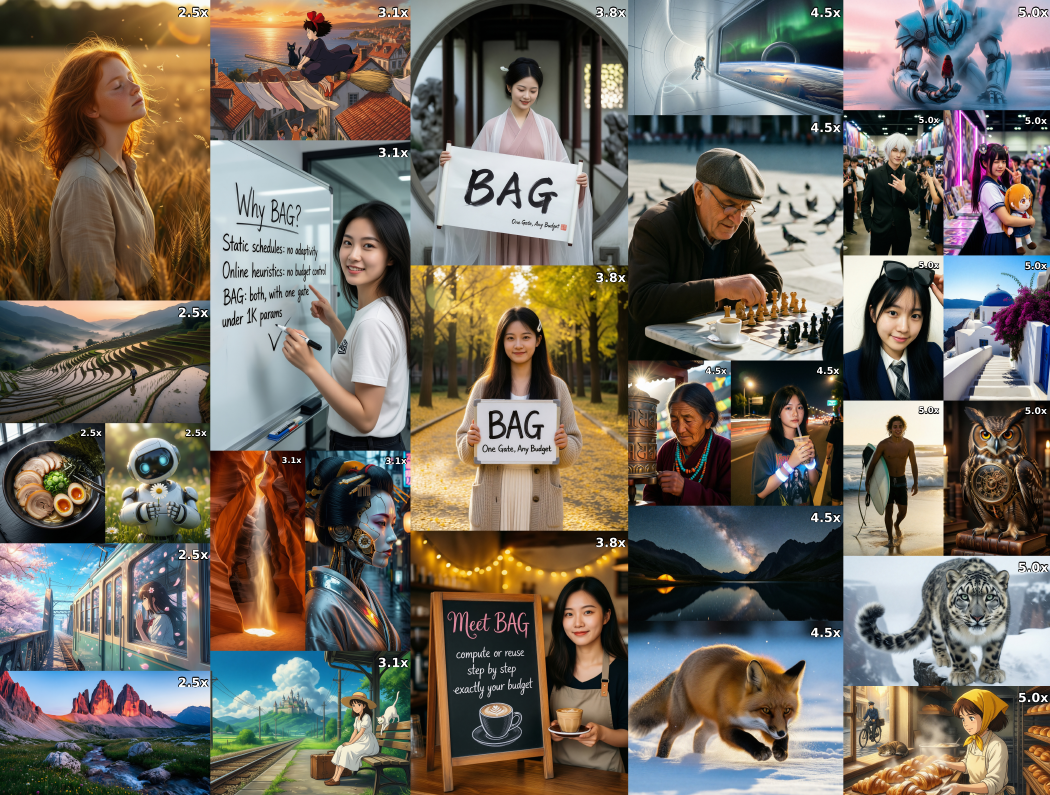}
\caption{\textbf{One gate, any budget.} Samples from Qwen-Image-2512~\citep{qwenimage} accelerated by a single \bag{} gate (under 1K parameters), with the speedup increasing from 2.5$\times$ on the left to 5.0$\times$ on the right; the label at the top-right corner of each image gives its speedup. The same gate serves every budget, prompt, and resolution shown, and spends exactly the requested compute budget.}
\label{fig:teaser}
\end{figure*}

Diffusion Transformers (DiTs) \citep{peebles2023dit} have established themselves as the standard architecture for state-of-the-art text-to-image and text-to-video generation \citep{flux2024,wan2025,hunyuanvideo,longcatvideo,zimage,qwenimage}. However, generating a single sample requires dozens of sequential forward passes through the network \citep{ho2020ddpm,lipman2022flow}, leading to high latency and substantial serving costs. To mitigate this computational bottleneck, several acceleration paradigms have been explored, including fast numerical solvers \citep{lu2022dpmsolver,zhao2023unipc,zhu2025distilling,wang2026adaptive}, model distillation into few-step or even single-step generators \citep{salimans2022progressive,song2023consistency}, and post-training quantization \citep{qdiff,qdit}. Among these approaches, diffusion caching \citep{ma2024deepcache,selvaraju2024fora,liu2025teacache,cachemeifyoucan,dydit,cachequant,blockdance} offers a particularly lightweight and complementary strategy. Because internal feature maps evolve gradually between adjacent denoising steps, certain steps can bypass full network evaluation by reusing cached computations. Caching requires no architecture modifications, retraining, or alterations to the underlying sampler, making it orthogonal to other acceleration techniques.
Given a $T$-step sampling process and a budget $B$ representing the total number of allowed function evaluations (NFEs), a caching policy must decide at each step whether to perform a full computation or reuse cached features. These $T$ sequential binary decisions form a cache schedule, which directly governs the trade-off between inference speed and generation quality.

\begin{figure*}[t]
\centering
\includegraphics[width=0.97\textwidth]{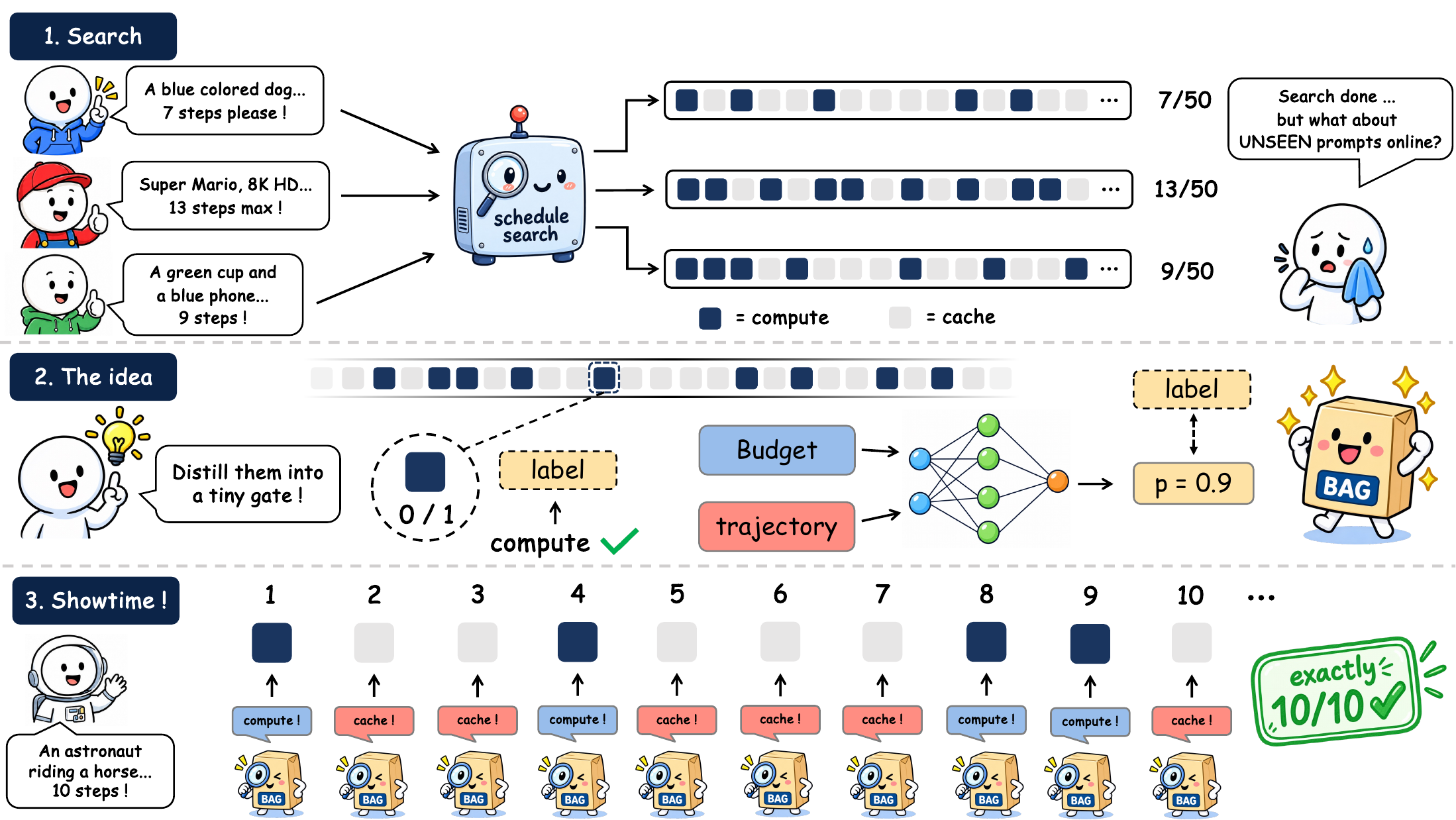}
\caption{\textbf{\bag{} overview: learn the scheduler, not the schedule.} Among the compared scheduling paradigms, \bag{} deploys the only per-step rule that reads both the remaining budget and the realized trajectory. (1)~\emph{Search} (training only, once per backbone): for each (prompt, budget) cell, a matched-NFE search produces a reference schedule: a full-horizon allocation of exactly $B$ evaluations, optimized to preserve the full-compute output. (2)~\emph{Distill}: each reference rollout is decomposed into per-step decision examples, from which a tiny gate learns to map budget state and trajectory feedback to compute-or-reuse decisions. (3)~\emph{Deploy}: on an unseen prompt the frozen gate decides closed-loop at every step, spending exactly the requested budget.}
\label{fig:method}
\end{figure*}

Existing diffusion caching methods fall into two main paradigms. The first paradigm uses online heuristic rules based on proxy signals like feature drift \citep{liu2025teacache,kahatapitiya2024adacache,chung2026seacache}. While instance-adaptive and lightweight, these rules lack budget awareness: they ignore the remaining budget when making per-step decisions, preventing global computation pacing. In addition, the realized NFE can only be controlled indirectly by tuning the threshold. Conversely, the second paradigm relies on static offline schedules, using predefined intervals \citep{ma2024deepcache,selvaraju2024fora,lv2024fastercache} or search-based timetables \citep{ma2024l2c,lei2026budcache}. Although static plans guarantee exact budget adherence via global planning, their open-loop nature imposes identical schedules on all instances and requires re-searching for every new budget constraint. Ultimately, neither paradigm combines real-time trajectory feedback with global budget pacing.

\begin{figure}[t]
\centering
\includegraphics[width=\columnwidth]{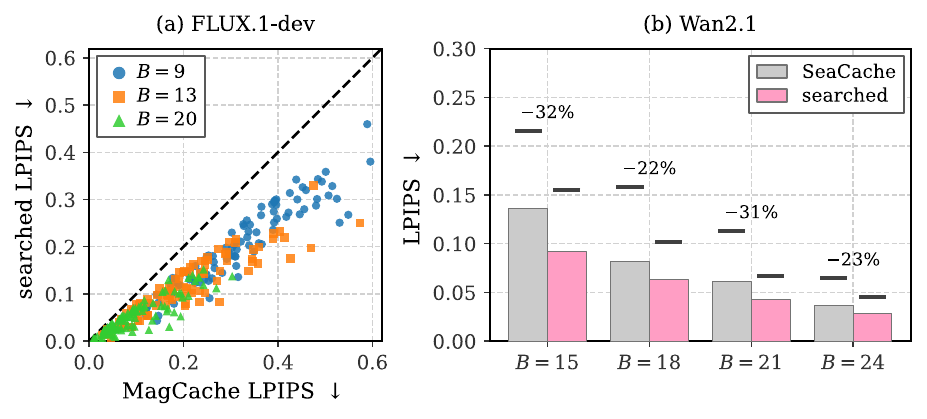}
\caption{\textbf{Both paradigms are suboptimal per prompt at matched NFE.} (a)~FLUX.1-dev, 100 prompts at three budgets: per-prompt reconstruction LPIPS of the searched schedule vs.\ MagCache at the same NFE; points below the dashed line favor the searched schedule. (b)~Wan2.1, 50 prompts: against SeaCache, the mean LPIPS to the full-compute output drops by 22--32\% (ticks: 90th percentile). At a given budget the deployed allocation is therefore not the best available, and how much room is left varies from prompt to prompt.}
\label{fig:headroom}
\end{figure}

To quantify the performance lost by current paradigms, \cref{fig:headroom} compares deployed methods against a reference schedule obtained via prompt-specific offline search~\citep{HC_opt,SA_opt} under identical NFE constraints. Across prompts and budgets, the prompt-specific schedule preserves full-compute generation quality substantially better than existing online and static baselines. Furthermore, both the offline-optimized budget allocation and the resulting quality gains vary substantially across different prompts. These results demonstrate the inherent suboptimality of existing paradigms: hand-crafted heuristics lack global budget awareness, static schedules lack instance adaptivity. However, prompt-specific search is computationally prohibitive for real-time inference.

To bridge this gap, an effective cache policy must simultaneously monitor two key signals at every step: the budget state, which determines the overall pacing of remaining compute, and the local trajectory, which evaluates whether the current step warrants full computation. Rather than hand-crafting complex decision heuristics, we propose learning this policy directly. Our method, \bag{} (\texttt{B}udget-\texttt{A}ware \texttt{G}ating), introduces a lightweight gating network of under 1K parameters that conditions its decisions on two distinct sources of information: global budget constraints and local trajectory dynamics.
Specifically, the global budget context is captured by three scalar indicators: the target compute ratio, the fraction of budget remaining, and the relative budget tightness over the remaining horizon. Simultaneously, local trajectory dynamics are reflected by three fine-grained signals: the elapsed steps since the previous full computation, the relative feature drift accumulated since that computation, and the step-to-step feature variation. From these six scalar inputs, the gate outputs a binary compute-or-reuse decision at a cost negligible relative to a network forward pass.
This dual-context design addresses the shortcomings of prior paradigms: trajectory feedback enables fine-grained, per-step adaptation, explicit budget conditioning coordinates computation over the remaining horizon, and a learned decision rule replaces hand-crafted, setting-specific heuristics.

To train the policy network, we formulate the learning process as offline-to-online schedule distillation. During training, we perform offline searches across diverse prompts and target budgets to construct high-quality reference schedules. Because each reference schedule is optimized with full access to the sampling trajectory and global budget, it serves as a strong target. We then replay these rollouts, extract the corresponding budget and trajectory states at each step, and train the gate via supervised learning to replicate the reference decisions. Consequently, global horizon planning is performed once offline, while only the compact learned policy runs during inference. By training across a range of target budgets, the gate learns how offline-optimized compute allocation shifts as a function of available budget $B$, effectively learning a generalized scheduling policy rather than a fixed schedule. At inference, the frozen gate makes closed-loop decisions on unseen prompts. The target budget $B$ is set directly at runtime and met exactly by construction. A single checkpoint therefore supports different budgets and step counts without retraining or re-search, while remaining robust to changes in seed, resolution, and guidance scale.

We treat high-quality cache schedules across prompts and budgets as outputs of one decision rule that reads both the remaining budget and the realized trajectory, and propose \bag{}, an offline-to-online schedule distillation framework that learns a lightweight budget- and trajectory-conditioned gate from searched reference schedules. Once trained, the same gate supports all evaluated budgets and sampling step counts, spends exactly the requested number of NFEs, and requires no further search or training. Comprehensive experiments on FLUX.1-dev~\citep{flux2024}, Wan2.1~\citep{wan2025}, and Qwen-Image-2512~\citep{qwenimage} demonstrate consistent improvements over state-of-the-art static and online caching methods across all reconstruction metrics under matched computation, including a ${+}1.2$\,dB PSNR gain over the strongest baseline at the $\sim5\times$ FLUX acceleration tier.

\section{Related Work}

\label{sec:related}

\paragraph{Cache scheduling.} \looseness=-1 Existing schedulers fall into three groups by how the schedule is obtained. \emph{Open-loop static schedules} fix reuse patterns before sampling: fixed-interval reuse in U-Nets~\citep{ma2024deepcache} and DiTs~\citep{selvaraju2024fora,chen2024deltadit}, attention-level broadcast~\citep{zhao2024pab}, and fixed timetables with CFG-branch reconstruction~\citep{lv2024fastercache}. MagCache~\citep{ma2025magcache} belongs here in effect: its residual-magnitude ratios come from an offline calibration curve rather than the running sample, so every prompt receives the same schedule. \emph{Online threshold rules}~\citep{prompttea,dicache,sencache} decide per step from a cheap trajectory signal: TeaCache~\citep{liu2025teacache} thresholds a rescaled input change, SeaCache~\citep{chung2026seacache} a spectrally filtered distance, and AdaCache~\citep{kahatapitiya2024adacache} maps a block-residual distance through a codebook of cache rates. \emph{Learned or optimized schedules}~\citep{dpcache,lemica,ertacache,meancache} obtain the schedule from data: Learning-to-Cache~\citep{ma2024l2c} learns an input-invariant layer routing, BudCache~\citep{lei2026budcache} searches a step schedule under a fixed budget with simulated annealing, and, concurrently with our work, ReCache~\citep{aliev2026recache} learns a budget-conditioned distribution over offline schedules with REINFORCE; what they deploy, however, remains a static per-budget schedule, identical for every prompt (for ReCache, a deterministic top-$k$ selection per budget). We regard ReCache and \bag{} as concurrent explorations of learned cache scheduling. \bag{} differs from all three groups in what is deployed: an online gate conditioned on budget state and realized trajectory feedback, supervised by offline-searched references; \cref{tab:paradigm} (\cref{app:baselines}) summarizes the paradigms.

\paragraph{Cache reuse mechanisms.} Orthogonal to \emph{when} to compute, a second axis changes what cached steps do: residual reuse~\citep{chen2024deltadit}, Taylor forecasting~\citep{liu2025taylorseer}, token-selective refresh~\citep{zou2024toca}, and frequency-domain CFG reconstruction~\citep{lv2024fastercache}. \bag{} fixes the mechanism to plain residual reuse and changes only the schedule, so the two axes stay decoupled and mechanism-level improvements can still be combined with it; TaylorSeer serves as a matched-NFE reference point on this axis.

\paragraph{Beyond caching.}
Diffusion inference is also accelerated by fast solvers, which reduce the number of denoising steps through higher-order integration of the reverse-time ODE~\citep{lu2022dpmsolver,zhao2023unipc,zhao2026dyweight,Yuan_2026_CVPR}; by distillation into few-step generators~\citep{salimans2022progressive,song2023consistency,hypersd,dmd,dmd2}, at the cost of an additional training stage; and by quantization~\citep{qdiff,qdit}, which lowers the cost of each network evaluation. These routes modify the sampler or the network itself, whereas caching removes redundant computation from a fixed sampler; the two levels compose naturally, and our Wan2.1 experiments cache a UniPC sampler.

\section{Method}
\label{sec:method}

\subsection{Preliminaries}

\noindent\textbf{Problem setup.}
A sampler runs $T$ denoising steps ($T{=}50$ for training and the main comparison). A \emph{cache schedule} is a binary mask $\mathbf{m} = (m_0, \ldots, m_{T-1})$, $m_t \in \{0,1\}$: step $t$ runs the network and refreshes the cache if $m_t{=}1$, and otherwise reuses the cached computation through the standard residual path, which we hold fixed for every scheduling method studied. The \emph{budget} fixes the number of function evaluations (NFEs) to $\|\mathbf{m}\|_1 = B$, with $m_0{=}1$ forced. Writing $\mathbf{x}_T(\mathbf{m},p)$ for the output on prompt $p$ under schedule $\mathbf{m}$, caching seeks to approximate the same-prompt, same-seed full-compute output $\mathbf{x}_T(\mathbf{1},p)$ with only $B$ of the $T$ evaluations. We evaluate reconstruction with PSNR, SSIM~\citep{ssim}, and LPIPS~\citep{zhang2018lpips}; some works instead report reference-free scores such as ImageReward~\citep{xu2023imagereward}. We compare all methods on the three reconstruction metrics.

\vspace{0.7em}
\noindent\textbf{Two scheduling paradigms.}
\looseness=-1
Static methods fix the full schedule before sampling, $\mathbf{m} = \boldsymbol{\mu}(B,T)$, based only on the budget and step count. They meet the budget exactly through full-horizon allocation, but use the same open-loop plan for every prompt and require a new plan for each budget. Online threshold rules decide per step, $m_t = \mathbb{I}[\,q_t > \delta\,]$, where $q_t$ is a hand-crafted trajectory signal: decisions respond to the realized rollout, but the realized computation is only an indirect consequence of $\delta$, and neither the remaining budget nor the horizon enters the decision.

\vspace{0.7em}
\noindent\textbf{Motivation.}
\cref{fig:headroom} measures what these restrictions cost: at matched NFE, per-prompt searched schedules reconstruct the full-compute output markedly better than what either paradigm deploys (MagCache~\citep{ma2025magcache}, static in effect, on FLUX; SeaCache~\citep{chung2026seacache} on Wan), and the experiments quantify the margins. The search itself is far too expensive to run per prompt, but its product, a full-horizon allocation of the budget, can be obtained once, offline, and distilled into a policy cheap enough to consult at every step. This is what \bag{} does: it keeps the per-step decision form of the online paradigm and gives it the budget awareness of the offline one, as a learned rule over an explicit state,
\begin{equation}
z_t = g_\theta\big(\mathbf{s}_t^{\mathrm{bud}},\, \mathbf{s}_t^{\mathrm{traj}}\big),
\label{eq:gate}
\end{equation}
whose two halves supply the two missing ingredients: global resource context and local rollout context.

\subsection{\bag{}: Budget-Aware Gating}
\bag{} has three stages (\cref{fig:method}): offline reference search, supervised distillation, and budget-exact inference.

\vspace{0.7em}
\noindent\textbf{Offline references.}
To supervise this rule, for each training prompt $p$ and budget $B$, a matched-NFE local search finds the schedule that best preserves the full-compute output under exactly $B$ evaluations, the \emph{searched reference schedule}
\begin{equation}
\hat{\mathbf{m}}(p,B) \,\approx\, \arg\min_{\|\mathbf{m}\|_1 = B}\; D\big(\mathbf{x}_T(\mathbf{m},p),\, \mathbf{x}_T(\mathbf{1},p)\big),
\label{eq:ref}
\end{equation}
where $D$ is the LPIPS distance. The search runs once per backbone, at training time only; it holds $B$ fixed and optimizes only where computation is placed, so its objective is aligned with full-compute fidelity. Because one gate later serves every budget, each training prompt is searched at each training budget, so the examples record how the allocation shifts as $B$ changes. Optimizer details and cost are given in \cref{app:search}.
\begin{algorithm}[t]
\caption{\bag{} training (one backbone, run once)}
\label{alg:train}
\begin{algorithmic}[1]
\STATE \textbf{Given:} prompts $\mathcal{P}$, budgets $\mathcal{B}$, steps $T$
\STATE \textbf{Initialize:} example set $\mathcal{D} \gets \emptyset$
\FOR{$p \in \mathcal{P}$}
    \STATE run the full-compute rollout \hfill $\triangleright~\textcolor{commentcolor}{\text{target }\mathbf{x}_T(\mathbf{1},p)}$
    \FOR{$B \in \mathcal{B}$}
        \STATE $\hat{\mathbf{m}} \gets$ matched-NFE search \hfill $\triangleright~\textcolor{commentcolor}{\text{Eq.~\eqref{eq:ref}, offline}}$
        \STATE save $\mathbf{s}_t^{\mathrm{bud}},\mathbf{s}_t^{\mathrm{traj}}$ along its last rollout \hfill $\triangleright~\textcolor{commentcolor}{\text{Eq.~\eqref{eq:state}}}$
        \STATE $\mathcal{D} \gets \mathcal{D} \cup \{(\mathbf{s}_t^{\mathrm{bud}}, \mathbf{s}_t^{\mathrm{traj}}, \hat{m}_t)\}_{t \geq 1}$
    \ENDFOR
\ENDFOR
\STATE z-score inputs; re-weight the positive class
\STATE fit $\theta$ on $\mathcal{D}$ \hfill $\triangleright~\textcolor{commentcolor}{\text{Eq.~\eqref{eq:loss}, ${\approx}1$ min}}$
\STATE \textbf{Return:} gate $g_\theta$
\end{algorithmic}
\end{algorithm}
\begin{algorithm}[t]
\caption{\bag{} budget-exact inference (one prompt)}
\label{alg:gate}
\begin{algorithmic}[1]
\STATE \textbf{Given:} gate $g_\theta$, budget $B$, steps $T$, cutoff $\tau{=}0.5$
\STATE \textbf{Initialize:} $m_0 \gets 1$; computed count $c \gets 1$
\FOR{$t = 1$ \TO $T-1$}
    \STATE read $\mathbf{s}_t^{\mathrm{bud}},\, \mathbf{s}_t^{\mathrm{traj}}$ \hfill $\triangleright~\textcolor{commentcolor}{\text{cheap online state, Eq.~\eqref{eq:state}}}$
    \STATE $m_t \gets \mathbb{I}\big[\sigma(g_\theta(\mathbf{s}_t^{\mathrm{bud}}, \mathbf{s}_t^{\mathrm{traj}})) > \tau\big]$ \hfill $\triangleright~\textcolor{commentcolor}{\text{gate decision}}$
    \STATE \textbf{if} $B{-}c \geq T{-}t$\textbf{:}\, $m_t \gets 1$ \hfill $\triangleright~\textcolor{commentcolor}{\text{spend remainder}}$
    \STATE \textbf{if} $B{-}c = 0$\textbf{:}\, $m_t \gets 0$ \hfill $\triangleright~\textcolor{commentcolor}{\text{budget exhausted}}$
    \STATE \textbf{if} $m_t{=}1$\textbf{:}\, run network; refresh cache; $c \gets c{+}1$
    \STATE \textbf{else:}\, reuse the cached residual
\ENDFOR
\STATE \textbf{Return:} final sample \hfill $\triangleright~\textcolor{commentcolor}{\text{realized NFE}=B\text{ exactly}}$
\end{algorithmic}
\end{algorithm}

\vspace{0.7em}
\noindent\textbf{Budget and trajectory state.}
The gate's state instantiates the two contexts of Eq.~\eqref{eq:gate} with six scalars. At step $t$, with $c_t$ evaluations already spent, it reads
\begin{equation}
\begin{aligned}
\mathbf{s}_t^{\mathrm{bud}} &= \Big[\,\tfrac{B}{T},\;\; \tfrac{B-c_t}{B},\;\; \tfrac{B-c_t}{T-t}\,\Big],\\[2pt]
\mathbf{s}_t^{\mathrm{traj}} &= \Big[\,t - t_{\mathrm{last}},\;\; \tfrac{\|\mathbf{x}_t - \mathbf{x}_{t_{\mathrm{last}}}\|}{\|\mathbf{x}_{t_{\mathrm{last}}}\|},\;\; \tfrac{\|\mathbf{x}_t - \mathbf{x}_{t-1}\|}{\|\mathbf{x}_{t-1}\|}\,\Big],
\end{aligned}
\label{eq:state}
\end{equation}
where $t_{\mathrm{last}}$ is the last computed step and $\mathbf{x}_t$ the post-patch-embedding token tensor (conditional branch for CFG models), the same class of observable that prior heuristics threshold. The budget state expresses the overall compute ratio, the remaining resource fraction, and the budget pressure over the remaining horizon: it is what lets the gate pace spending globally, and, because its entries are ratios of step counts, the same definition applies at other budgets and step counts. The trajectory state carries cache staleness, cache drift, and the local step change: together they measure how far the rollout has moved since the last refresh and how fast it is moving now. Each scalar is a counter or a single reduction over a tensor the forward pass already produces, with no extra network evaluation, and the gate learns their joint effect.

\vspace{0.7em}
\noindent\textbf{Supervised distillation.}
\looseness=-1
The per-step supervision comes out of the search itself (\cref{alg:train}): its last round rolls out the returned $\hat{\mathbf{m}}$, and along this rollout we save the states at every step, yielding examples $\mathcal{D} = \big\{(\mathbf{s}_t^{\mathrm{bud}}, \mathbf{s}_t^{\mathrm{traj}}, \hat{m}_t)\big\}$ over all training prompts and budgets ($t \geq 1$). The gate is trained by per-step binary classification,
\begin{equation}
\mathcal{L}(\theta) = \textstyle\sum_{\mathcal{D}} \operatorname{BCE}\big(\sigma(z_t),\, \hat{m}_t\big),
\label{eq:loss}
\end{equation}
where $\sigma$ is the logistic sigmoid. The inputs are z-scored, and the BCE loss is class-balanced. The budget features supply the sequence context, so a sequence-level allocation problem is reduced to per-step classification, and the global count comes out exact at inference. The gate is a small MLP of under 1K parameters, and one gate is trained per backbone in about a minute on a single RTX~4090. Because deployment states are induced by the gate's own past decisions rather than the reference rollouts, distillation incurs an off-policy state-distribution shift. The reported results are measured under deployment and already include its cost; correcting the shift with on-policy relabeling would multiply the offline cost, so we leave it to future work. Architectures, optimizers, and further discussion are in \cref{app:gate}.
\begin{table}[t]
\centering
\footnotesize
\tabcolsep=5.1pt
\begin{tabular}{l c c c c c c}
\toprule
Method & NFE & Lat.\,(s) & Speed & PSNR$\uparrow$ & SSIM$\uparrow$ & LPIPS$\downarrow$ \\
\midrule
FLUX.1-dev & 50 & 31.13 & 1.00$\times$ & -- & -- & -- \\
\midrule
\multicolumn{7}{l}{\emph{$\sim$5$\times$ acceleration ($B{=}9$)}} \\
\dashedmidrule{7}
10 steps & 10 & 6.23 & 4.99$\times$ & 14.73 & 0.632 & 0.4823 \\
TeaCache & 10.6 & 6.85 & 4.54$\times$ & 16.09 & 0.672 & 0.4266 \\
MagCache & 10.0 & 6.25 & 4.98$\times$ & \underline{20.03} & 0.745 & \underline{0.3014} \\
TaylorSeer & 9.0 & 7.02 & 4.43$\times$ & 15.44 & 0.643 & 0.4399 \\
BudCache$^\dagger$ & 9.0 & 5.63 & 5.53$\times$ & 19.49 & 0.727 & 0.3275 \\
SeaCache & 9.0 & 6.00 & 5.19$\times$ & 19.66 & \underline{0.755} & 0.3031 \\
\rowcolor{lightCyan}\bag{} (ours) & 9.0 & 5.99 & 5.20$\times$ & \textbf{21.27} & \textbf{0.766} & \textbf{0.2887} \\
\midrule
\multicolumn{7}{l}{\emph{$\sim$3.8$\times$ acceleration ($B{=}13$)}} \\
\dashedmidrule{7}
15 steps & 15 & 9.33 & 3.34$\times$ & 15.77 & 0.673 & 0.4118 \\
TeaCache & 14.7 & 9.38 & 3.32$\times$ & 17.49 & 0.724 & 0.3395 \\
MagCache & 13.0 & 8.11 & 3.84$\times$ & 21.70 & 0.812 & 0.2111 \\
TaylorSeer & 14.0 & 10.01 & 3.11$\times$ & 18.29 & 0.744 & 0.2841 \\
BudCache$^\dagger$ & 13.0 & 8.11 & 3.84$\times$ & 21.55 & 0.805 & 0.2136 \\
SeaCache & 13.0 & 8.46 & 3.68$\times$ & \underline{21.85} & \underline{0.821} & \underline{0.2030} \\
\rowcolor{lightCyan}\bag{} (ours) & 13.0 & 8.48 & 3.67$\times$ & \textbf{24.46} & \textbf{0.849} & \textbf{0.1675} \\
\midrule
\multicolumn{7}{l}{\emph{$\sim$2.4$\times$ acceleration ($B{=}20$)}} \\
\dashedmidrule{7}
25 steps & 25 & 15.49 & 2.01$\times$ & 18.10 & 0.753 & 0.2929 \\
TeaCache & 21.0 & 13.28 & 2.34$\times$ & 18.84 & 0.765 & 0.2744 \\
MagCache & 20.0 & 12.47 & 2.50$\times$ & 25.74 & 0.892 & 0.1105 \\
TaylorSeer & 26.0 & 17.14 & 1.82$\times$ & 23.50 & 0.870 & 0.1325 \\
BudCache$^\dagger$ & 20.0 & 12.46 & 2.50$\times$ & 27.52 & 0.903 & 0.0898 \\
SeaCache & 20.9 & 13.37 & 2.33$\times$ & \underline{27.81} & \underline{0.914} & \underline{0.0835} \\
\rowcolor{lightCyan}\bag{} (ours) & 20.0 & 12.81 & 2.43$\times$ & \textbf{29.18} & \textbf{0.918} & \textbf{0.0773} \\
\bottomrule
\end{tabular}
\caption{\textbf{Quantitative results on FLUX.1-dev.} One \bag{} checkpoint improves all three metrics at every acceleration tier. Methods are compared at matched NFE (baselines spend at least as many evaluations), with wall-clock latency also reported. \textbf{Bold}: best per tier; \underline{underline}: second; top row: full-compute reference. $^\dagger$BudCache under our search (see \cref{app:baselines}).}
\label{tab:flux}
\end{table}
\begin{table}[t]
\centering
\footnotesize
\tabcolsep=5.1pt
\begin{tabular}{l c c c c c c}
\toprule
Method & NFE & Lat.\,(s) & Speed & PSNR$\uparrow$ & SSIM$\uparrow$ & LPIPS$\downarrow$ \\
\midrule
Wan2.1-1.3B & 50 & 184.40 & 1.00$\times$ & -- & -- & -- \\
\midrule
\multicolumn{7}{l}{\emph{$\sim$3.4$\times$ acceleration ($B{=}15$)}} \\
\dashedmidrule{7}
TeaCache & 17.0 & 62.95 & 2.93$\times$ & 21.01 & 0.769 & 0.1827 \\
MagCache & 15.0 & 55.58 & 3.32$\times$ & 20.31 & 0.746 & 0.2047 \\
BudCache$^\dagger$ & 15.0 & 55.58 & 3.32$\times$ & \underline{24.06} & \underline{0.835} & \underline{0.1241} \\
SeaCache & 15.0 & 56.70 & 3.25$\times$ & 22.99 & 0.805 & 0.1470 \\
\rowcolor{lightCyan}\bag{} (ours) & 15.0 & 55.76 & 3.31$\times$ & \textbf{24.96} & \textbf{0.848} & \textbf{0.1142} \\
\midrule
\multicolumn{7}{l}{\emph{$\sim$2.7$\times$ acceleration ($B{=}19$)}} \\
\dashedmidrule{7}
TeaCache & 20.0 & 74.22 & 2.48$\times$ & 23.01 & 0.826 & 0.1292 \\
MagCache & 19.0 & 70.57 & 2.61$\times$ & 25.71 & 0.878 & 0.0891 \\
BudCache$^\dagger$ & 19.0 & 70.47 & 2.62$\times$ & 24.93 & 0.861 & 0.1010 \\
SeaCache & 19.0 & 72.29 & 2.55$\times$ & \underline{26.28} & \underline{0.880} & \underline{0.0871} \\
\rowcolor{lightCyan}\bag{} (ours) & 19.0 & 70.68 & 2.61$\times$ & \textbf{27.89} & \textbf{0.905} & \textbf{0.0685} \\
\midrule
\multicolumn{7}{l}{\emph{$\sim$2.1$\times$ acceleration ($B{=}24$)}} \\
\dashedmidrule{7}
TeaCache & 25.0 & 93.09 & 1.98$\times$ & 23.79 & 0.846 & 0.1110 \\
MagCache & 26.0 & 96.77 & 1.91$\times$ & 26.96 & 0.906 & 0.0639 \\
BudCache$^\dagger$ & 24.0 & 89.38 & 2.06$\times$ & 26.07 & 0.886 & 0.0802 \\
SeaCache & 24.6 & 92.30 & 2.00$\times$ & \underline{30.72} & \underline{0.943} & \underline{0.0395} \\
\rowcolor{lightCyan}\bag{} (ours) & 24.0 & 89.55 & 2.06$\times$ & \textbf{31.38} & \textbf{0.947} & \textbf{0.0381} \\
\bottomrule
\end{tabular}
\caption{\textbf{Quantitative results on Wan2.1-T2V-1.3B.} One \bag{} checkpoint improves all three metrics at every acceleration tier. Methods are compared at matched NFE (baselines spend at least as many evaluations), with per-video wall-clock latency also reported. \textbf{Bold}: best per tier; \underline{underline}: second; top row: full-compute reference. $^\dagger$BudCache under our search (see \cref{app:baselines}).}
\label{tab:wan}
\end{table}

\vspace{0.7em}
\noindent\textbf{Budget-exact inference.}
\looseness=-1
At deployment the gate runs inside the sampling loop: whenever $\sigma(z_t) > \tau$ (a fixed cutoff, $\tau{=}0.5$) the step computes and one evaluation is spent, and the updated budget state enters the next step's decision. Unlike the thresholds of prior heuristics, $\tau$ does not set how much is computed: the total is pinned by $B$ itself. A step with the budget already spent can only reuse, and a rollout with exactly as many steps left as budget can only compute them all; at such steps there is nothing to decide, and at every other step the gate's decision stands (\cref{alg:gate}). Outside the forced cases, the gate places evaluations from the budget and trajectory state. In the main comparisons, it spends all $B$ before the remainder rule activates and leaves a brief reuse tail, as do most reference schedules. The boundary rules ensure an NFE of $B$ without prescribing those placements, so the requested budget directly sets the speedup.

\section{Experiments}
\label{sec:exp}

\subsection{Setup}
\label{sec:exp:setup}

\noindent\textbf{Backbones and data.} We evaluate on three backbones. On FLUX.1-dev~\citep{flux2024} (50 steps, $1024^2$), we train the gate on 120 GenEval prompts~\citep{ghosh2023geneval} (96 train, 24 validation), searching each prompt at every budget in $\{7,10,13,16,19,22,25\}$, and test on all 200 DrawBench prompts~\citep{saharia2022imagen}, a disjoint prompt source. On Wan2.1-T2V-1.3B~\citep{wan2025} (50-step UniPC~\citep{zhao2023unipc}, $832{\times}480$, 65 frames), we train on 50 VBench~\citep{huang2024vbench} prompts, each searched at $\{15,18,21,24\}$, and test on 100 disjoint VBench prompts sampled uniformly over all 19 categories. On Qwen-Image-2512~\citep{qwenimage} (50 steps, $1024^2$), the FLUX protocol is reused unchanged: the gate is trained on schedules searched for the same 120 GenEval prompts and tested on DrawBench-200. The gate is also evaluated at budgets it was not trained on: $B{=}9$ and $B{=}20$ on FLUX, $B{=}19$ on Wan, and both Qwen-Image budgets. Every \bag{} number is thus reported on held-out prompts; the prompt-isolation protocol is in \cref{app:isolation}.

\begin{figure*}[t]
\centering
\includegraphics[width=\textwidth]{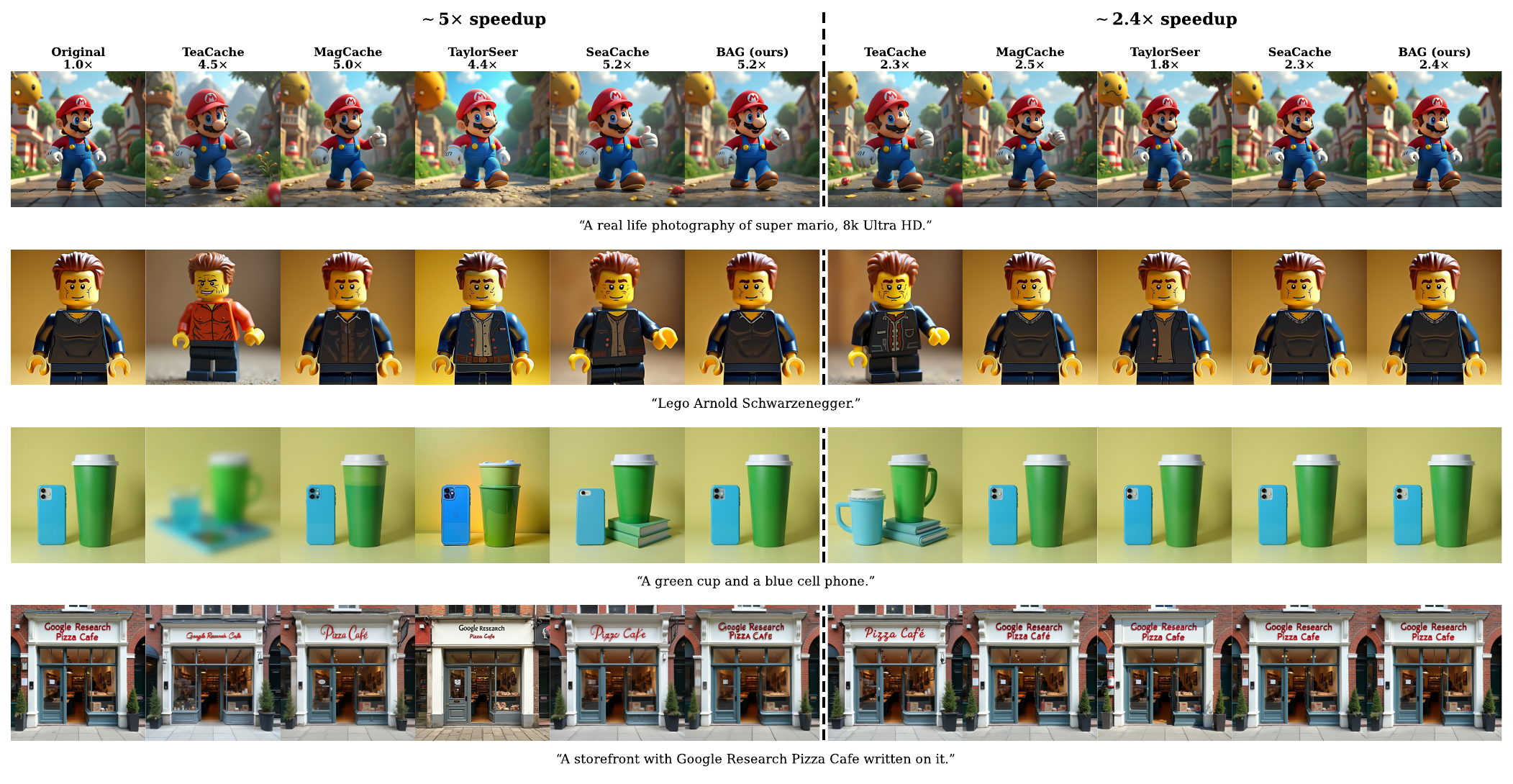}
\caption{\textbf{Qualitative results on FLUX.1-dev} at the $\sim$5$\times$ ($B{=}9$) and $\sim$2.4$\times$ ($B{=}20$) tiers, with the 50-step original leftmost. At matched compute the accelerated baselines drift in composition and object identity, while \bag{} stays closer to the full-compute output. More qualitative results in \cref{app:qual}.}
\label{fig:qualflux}
\end{figure*}
\begin{figure}[t]
\centering
\includegraphics[width=\columnwidth]{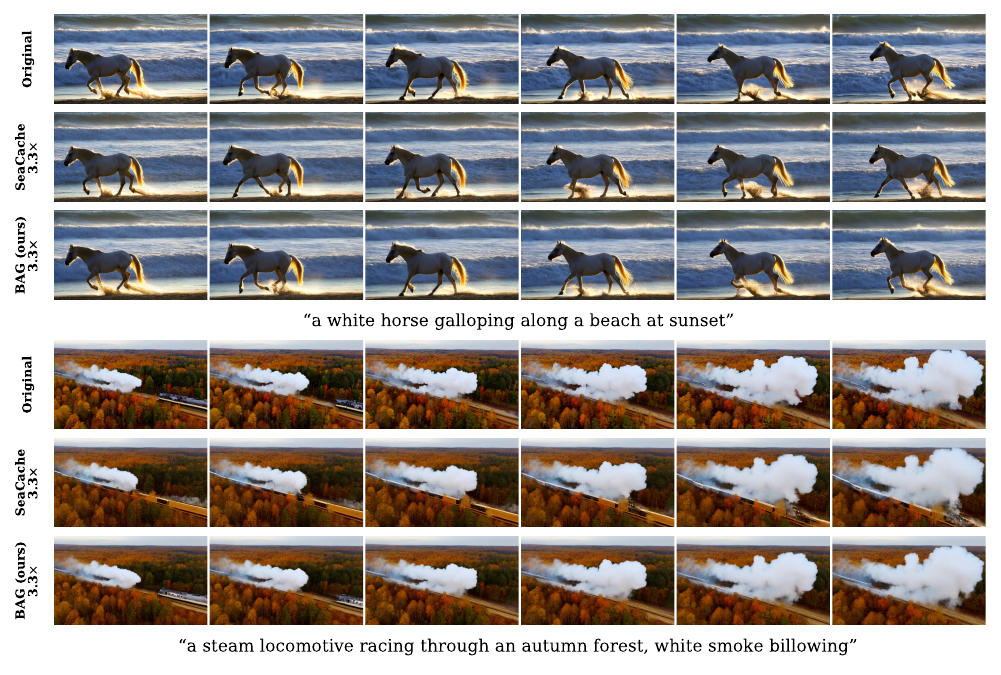}
\caption{\textbf{Qualitative results on Wan2.1} at the $\sim$3.4$\times$ tier ($B{=}15$, six frames). At matched NFE \bag{} follows the full-compute motion (the horse's gait, the locomotive's smoke) while SeaCache blurs fast motion and drifts. More qualitative results in \cref{app:qual}.}
\label{fig:qual}
\end{figure}
\begin{table}[t]
\centering
\footnotesize
\tabcolsep=5.1pt
\begin{tabular}{l c c c c c}
\toprule
Method & NFE & Speed & PSNR$\uparrow$ & SSIM$\uparrow$ & LPIPS$\downarrow$ \\
\midrule
Qwen-Image-2512 & 50 & 1.00$\times$ & -- & -- & -- \\
\midrule
\multicolumn{6}{l}{\emph{$\sim$5$\times$ acceleration ($B{=}9$)}} \\
\dashedmidrule{6}
9 steps & 9.0 & 5.56$\times$ & 14.05 & 0.549 & 0.4603 \\
SeaCache & 9.0 & 5.56$\times$ & 16.70 & 0.579 & 0.4314 \\
BudCache & 9.0 & 5.56$\times$ & \underline{18.76} & \underline{0.622} & \underline{0.3543} \\
\rowcolor{lightCyan}\bag{} (ours) & 9.0 & 5.56$\times$ & \textbf{19.13} & \textbf{0.629} & \textbf{0.3120} \\
\midrule
\multicolumn{6}{l}{\emph{$\sim$3.2$\times$ acceleration ($B{=}15$)}} \\
\dashedmidrule{6}
15 steps & 15.0 & 3.33$\times$ & 16.03 & 0.657 & 0.3427 \\
SeaCache & 15.0 & 3.33$\times$ & 20.60 & 0.769 & 0.2231 \\
BudCache & 15.0 & 3.33$\times$ & \underline{21.62} & \underline{0.801} & \underline{0.1804} \\
\rowcolor{lightCyan}\bag{} (ours) & 15.0 & 3.33$\times$ & \textbf{23.66} & \textbf{0.801} & \textbf{0.1759} \\
\bottomrule
\end{tabular}
\caption{\textbf{Quantitative results on Qwen-Image-2512.} One \bag{}
gate leads all three metrics at both acceleration tiers, with methods
compared at matched NFE. \textbf{Bold}: best per tier;
\underline{underline}: second; top row: full-compute reference.}
\label{tab:qwen}
\end{table}

\vspace{0.7em}
\noindent\textbf{Protocol.} All methods are evaluated against the same-seed, same-machine 50-step full-compute output of the same backbone, with per-prompt seeds. We report realized NFE (function evaluations per prompt), PSNR/SSIM/LPIPS, and latency. Baseline thresholds are swept \emph{on the test set} to match the target NFE, a protocol that favors the baselines. \bag{} spends exactly $B$ by construction, and unless stated otherwise every \bag{} number comes from one gate per backbone at $\tau{=}0.5$, all budgets served by the same checkpoint.

\vspace{0.7em}
\noindent\textbf{Baselines.} \looseness=-1 We compare against TeaCache~\citep{liu2025teacache}, MagCache~\citep{ma2025magcache}, SeaCache~\citep{chung2026seacache} (the strongest heuristic overall in our runs), TaylorSeer~\citep{liu2025taylorseer} (official implementation) as the representative of the mechanism axis, and, on FLUX and Qwen-Image, naive step reduction. Tiers match the $\sim$5/3.8/2.4$\times$ (FLUX) and $\sim$3.4/2.7/2.1$\times$ (Wan) acceleration factors targeted by prior caching work. For fairness, BudCache~\citep{lei2026budcache} uses our search on FLUX and Wan: we run the schedule search behind our gate's labels on BudCache's calibration prompt and broadcast the resulting schedule to all test prompts, isolating its one-prompt-calibration choice from search strength; on Qwen-Image it runs its official search protocol, since there the official search spends more rollouts per schedule than ours. Per-tier operating points and BudCache's official protocol are given in \cref{app:baselines}.

\subsection{Main Results}
\label{sec:exp:main}

\noindent\textbf{Quantitative comparison.}
\looseness=-2
\cref{tab:flux,tab:wan} present the matched-computation comparison against both paradigms. At every tested budget on both backbones, \bag{} improves all three reconstruction metrics over the strongest baseline using the same or fewer NFEs. The PSNR margin over SeaCache, the strongest online heuristic, reaches $+2.6$\,dB on FLUX and $+2.0$\,dB on Wan, and LPIPS is reduced by 21--22\% relative to SeaCache at the two tighter Wan tiers. Against the per-tier runner-up (MagCache at the tightest FLUX tier, BudCache$^\dagger$ at Wan $B{=}15$) the PSNR margin ranges from $+0.7$ to $+2.6$\,dB. Where SeaCache overshoots the tier, \bag{} wins while spending less. TaylorSeer sits on a different design axis, forecasting cached features rather than changing the schedule, and under this protocol it trails at every tier even when granted the next-higher point on its discrete NFE grid (\cref{app:baselines}). \cref{tab:vbenchmain} reports the reference-free VBench dimensions on Wan at the tightest tier, where every method stays within $0.003$ of the reference on the first four dimensions and \bag{} has the highest average among the accelerated methods. \cref{tab:qwen} extends the comparison to Qwen-Image-2512: the gate again leads every reconstruction metric at both tiers under matched computation, with PSNR margins of $+2.4$ and $+3.1$\,dB over SeaCache, and naive step reduction trails every caching method. VBench results for all three tiers, FLUX preference metrics, and further analyses are in \cref{app:tables}.

\begin{table*}[t]
\centering
\footnotesize
\renewcommand{\arraystretch}{1.1}
\tabcolsep=11pt
\begin{tabular}{l c c c c c c c c}
\toprule
Method & Subject$\uparrow$ & Background$\uparrow$ & Smoothness$\uparrow$ & Flickering$\uparrow$ & Dynamic$\uparrow$ & Aesthetic$\uparrow$ & Imaging$\uparrow$ & Avg$\uparrow$ \\
\midrule
Wan2.1 (50) & 0.9734 & 0.9742 & 0.9893 & 0.9825 & 0.3400 & 0.6133 & 0.7017 & 0.7963 \\
\midrule
TeaCache & 0.9730 & \underline{0.9726} & \textbf{0.9888} & \underline{0.9831} & \underline{0.3000} & \textbf{0.6065} & 0.6950 & \underline{0.7884} \\
MagCache & 0.9719 & \textbf{0.9732} & \underline{0.9887} & \textbf{0.9832} & 0.2800 & 0.6010 & 0.6881 & 0.7837 \\
BudCache$^\dagger$ & \underline{0.9738} & 0.9721 & 0.9886 & 0.9828 & \underline{0.3000} & 0.6041 & \textbf{0.6969} & 0.7883 \\
SeaCache & 0.9720 & 0.9719 & \textbf{0.9888} & 0.9828 & 0.2700 & 0.5991 & 0.6935 & 0.7826 \\
\rowcolor{lightCyan}\bag{} (ours) & \textbf{0.9743} & 0.9722 & \underline{0.9887} & 0.9828 & \textbf{0.3100} & \underline{0.6046} & \underline{0.6962} & \textbf{0.7898} \\
\bottomrule
\end{tabular}
\caption{\textbf{Reference-free VBench dimensions on Wan2.1} at the $\sim$3.4$\times$ tier ($B{=}15$; custom inputs). The other nine dimensions require the official prompt suite. All methods are within $0.003$ of the reference on the first four dimensions; \bag{} has the highest accelerated-method average. \textbf{Bold}/\underline{underline}: best/second-best accelerated method (ties share a marker). Top: full-compute reference; all tiers: \cref{tab:vbench}. $^\dagger$BudCache evaluated under our search protocol.}
\label{tab:vbenchmain}
\vspace{2ex}
\end{table*}

\begin{figure*}[t]
\centering
\captionsetup{skip=6pt}%
\includegraphics[width=\textwidth]{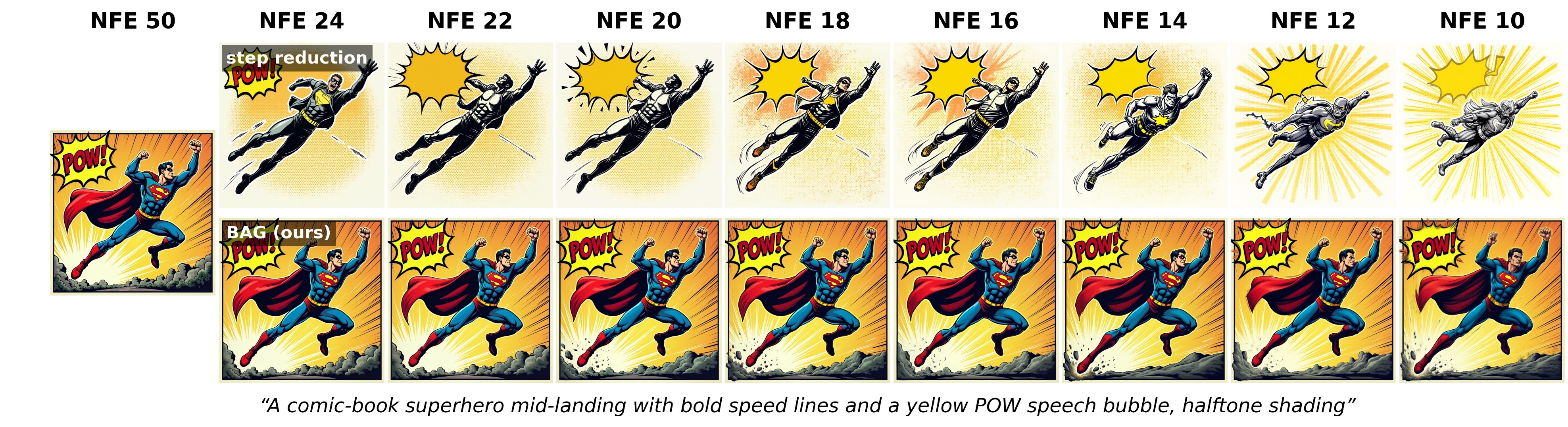}\\
\includegraphics[width=\textwidth]{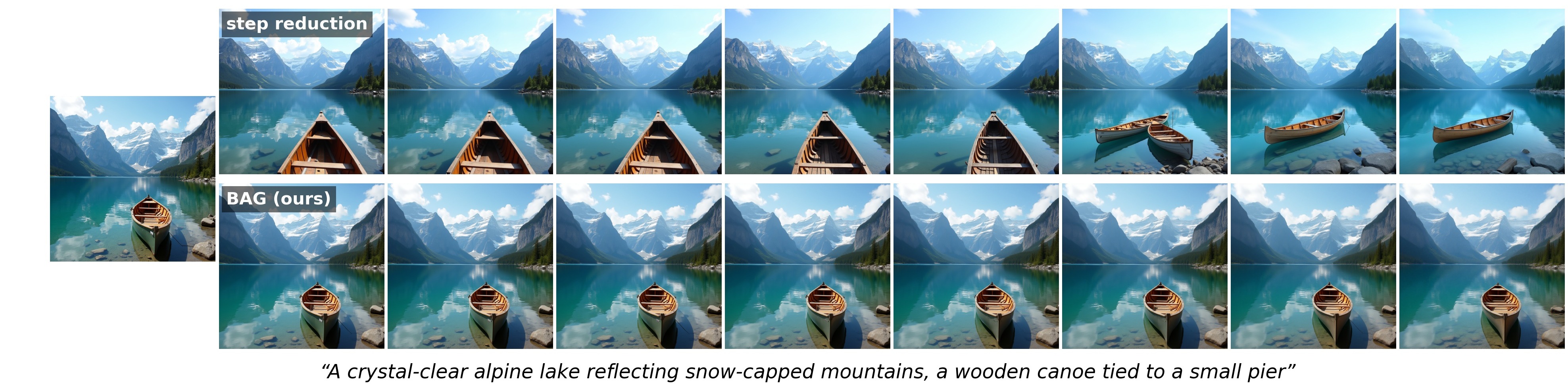}
\caption{\textbf{One \bag{} checkpoint across the budget range vs.\ naive step reduction} (two of the 50 sweep prompts; the full sweep is in \cref{fig:sweep,fig:sweep2}, metrics over all 50 in \cref{tab:sweep}). Leftmost: the same-seed 50-step full-compute output. Columns run from NFE 24 down to 10; per case, the top row is naive step reduction at that step count and the bottom row is \bag{} at budget $B{=}$NFE on the 50-step grid. Step reduction loses the \textsc{pow} bubble and the superhero's costume and settles on a different lake composition, while the single \bag{} gate tracks the full-compute output across the whole range.}
\label{fig:sweepmain}
\end{figure*}
\vspace{0.7em}
\noindent\textbf{Qualitative comparison.}
\cref{fig:qualflux} compares FLUX outputs at matched NFE: threshold schedules lose sign text, object identity, and scene layout, while \bag{} stays closer to the full-compute output. \cref{fig:qual} shows the corresponding Wan comparison at the tightest tier, and \cref{fig:teaser} shows Qwen-Image-2512 samples from the single gate across the whole $2.5$--$5\times$ range. Further cases and the realized schedules behind these comparisons appear in \cref{app:qual}.

\subsection{Ablations and Analysis}
\label{sec:exp:analysis}

\noindent\textbf{Reference supervision quality.}
We first verify that the searched references are worth distilling. \cref{fig:headroom} compares them, at matched NFE, with what each paradigm deploys: against MagCache on FLUX the searched schedules cut the mean LPIPS to the full-compute output by about $30\%$ at every budget, and against SeaCache on Wan2.1 by $22$--$32\%$. The references are thus better allocations than either deployed schedule, supporting their use as offline labels.

\vspace{0.7em}
\noindent\textbf{Ablation on the input state.}
\looseness=-1
\cref{tab:abl} removes each half of the state in Eq.~\eqref{eq:state} in turn, with the architecture, training data, and inference procedure unchanged: both variants still spend exactly $B$ evaluations, so what is tested is purely the allocation. Each half alone collapses to one of the existing paradigms. Without the trajectory state the gate reads only $(t, B, c_t)$ and deploys the same open-loop schedule for every prompt, a static method in effect; without the budget state it reacts only to the local trajectory, an online rule in effect. Neither is enough: across the three budgets of \cref{tab:abl-full}, the static-like variant falls behind even SeaCache ($0.5$--$1.5$\,dB PSNR) and the online-like variant loses $2.5$--$7.0$\,dB PSNR to the full gate. The two halves are complementary, and only reading both closes the paradigm gap (results at all budgets are in \cref{tab:abl-full}).

\vspace{0.7em}
\begin{table}[t]
\centering
\footnotesize
\renewcommand{\arraystretch}{1.1}
\tabcolsep=4pt
\begin{tabular}{lcccccc}
\toprule
 & \multicolumn{3}{c}{$B{=}9$} & \multicolumn{3}{c}{$B{=}13$} \\
\cmidrule(lr){2-4}\cmidrule(lr){5-7}
Variant & PSNR$\uparrow$ & SSIM$\uparrow$ & LPIPS$\downarrow$ & PSNR$\uparrow$ & SSIM$\uparrow$ & LPIPS$\downarrow$ \\
\midrule
SeaCache & 19.66 & 0.755 & 0.303 & 21.85 & 0.821 & 0.203 \\
\bag{} & \textbf{21.27} & \textbf{0.766} & \textbf{0.289} & \textbf{24.46} & \textbf{0.849} & \textbf{0.168} \\
w/o budget & 18.80 & 0.596 & 0.572 & 20.21 & 0.643 & 0.499 \\
w/o trajectory & 19.12 & 0.707 & 0.412 & 20.49 & 0.759 & 0.299 \\
\bottomrule
\end{tabular}
\caption{\textbf{Input-state ablation} ($B{=}9/13$; one gate per row). Each half of the state alone collapses the gate into one of the existing paradigms: budget-only deploys a single open-loop schedule like a static method, and trajectory-only reacts to local signals without pacing, like an online rule. Neither recovers the full gate, so the two contexts are complementary rather than redundant.}
\label{tab:abl}
\end{table}

\begin{table}[t]
\centering
\footnotesize
\renewcommand{\arraystretch}{1.1}
\tabcolsep=5.4pt
\begin{tabular}{llcccc}
\toprule
Deployment shift & Method & NFE & PSNR$\uparrow$ & SSIM$\uparrow$ & LPIPS$\downarrow$ \\
\midrule
\multirow{2}{*}{50-step (train-time)} & SeaCache & 13.0 & 21.85 & 0.821 & 0.203 \\
 & \bag{} & 13.0 & \textbf{24.46} & \textbf{0.849} & \textbf{0.168} \\
\dashedmidrule{6}
\multirow{2}{*}{25-step sampler} & SeaCache & 10.0 & 23.39 & 0.838 & 0.181 \\
 & \bag{} & 10.0 & \textbf{25.25} & \textbf{0.855} & \textbf{0.156} \\
\dashedmidrule{6}
\multirow{2}{*}{28-step sampler} & SeaCache & 10.6 & 23.18 & 0.834 & 0.189 \\
 & \bag{} & 10.0 & \textbf{24.44} & \textbf{0.845} & \textbf{0.175} \\
\midrule
\multirow{2}{*}{seed $42 \to 1042$} & SeaCache & 13.0 & 21.81 & 0.821 & 0.193 \\
 & \bag{} & 13.0 & \textbf{24.38} & \textbf{0.847} & \textbf{0.157} \\
\dashedmidrule{6}
\multirow{2}{*}{resolution $\to 768^2$} & SeaCache & 13.0 & 22.33 & 0.810 & 0.198 \\
 & \bag{} & 13.0 & \textbf{24.64} & \textbf{0.841} & \textbf{0.162} \\
\dashedmidrule{6}
\multirow{2}{*}{guidance $\to 5.0$} & SeaCache & 13.0 & 20.29 & 0.796 & 0.207 \\
 & \bag{} & 13.0 & \textbf{23.52} & \textbf{0.843} & \textbf{0.154} \\
\bottomrule
\end{tabular}
\caption{\textbf{One FLUX gate across sampler step counts and generation shifts.} Top: trained on 50-step trajectories and deployed unchanged on 25- and 28-step samplers at $B{=}10$; the 50-step setting is the anchor. Bottom: seed, resolution, and guidance shifts at $T{=}50$, $B{=}13$. Each row uses a full-compute reference at the \emph{same} setting and step count. SeaCache keeps $\delta{=}0.6$, re-verified per setting (realized NFE listed). \bag{} wins all metrics with no more NFEs.}
\label{tab:general}
\end{table}

\noindent\textbf{Robustness across deployment settings.}
\looseness=-1
A single checkpoint per backbone serves every evaluated budget, as \cref{tab:flux,tab:wan} show: the same weights produce every \bag{} number. \cref{tab:general} further tests this checkpoint under deployment shifts. Trained only on 50-step trajectories, it wins all three metrics using the same or fewer NFEs when deployed unchanged on 25- and 28-step samplers (top): the gate's inputs are ratios such as the remaining budget over the remaining steps, so the same checkpoint applies directly at other step counts. The margins persist under shifts of seed, resolution, and guidance on the 50-step sampler at $B{=}13$ (bottom), each scored against its own 50-step reference at the same setting. None of these settings is an input to the gate, and no per-setting re-tuning is involved. A finer budget sweep on further unseen prompts is reported in \cref{app:tables}; \cref{fig:sweepmain} shows two of these prompts, where naive step reduction loses the \textsc{pow} bubble and drifts in composition as the step count shrinks, while the single gate tracks the full-compute output down to $B{=}10$.

\subsection{Efficiency}
\label{sec:exp:latency}

\noindent\textbf{Deployment latency.}
\looseness=-1
Latency is NFE-linear on both backbones (Lat.\ columns of \cref{tab:flux,tab:wan}), measured on a single RTX~4090. Methods that fix the schedule offline make no decision at run time; online methods pay a small per-step cost. \bag{} decides at run time too, so it pays this cost as well. Measured against denoise time, on FLUX this amounts to ${\approx}3\%$ for TeaCache, ${\approx}5\%$ for \bag{} (two tensor norms and one pass through the decision MLP), and ${\approx}5$--$6\%$ for SeaCache, whose criterion computes a spectrally filtered (FFT-based) distance; on Wan, where each network step is far more expensive, the same decisions cost under $0.3\%$ for \bag{} and TeaCache and ${\approx}2\%$ for SeaCache. TaylorSeer's skipped steps are not free, since each evaluates a Taylor forecast of the cached features, and this cost is paid per skipped step while the compute it is measured against shrinks with the budget, so its relative overhead grows from ${\approx}6\%$ at NFE 26 to ${\approx}25\%$ at NFE 9 on FLUX. All comparisons are matched to each tier's target NFE, and \bag{} spends no more evaluations than any baseline.

\vspace{0.7em}
\noindent\textbf{Offline cost.}
\looseness=-1
The main one-time cost is offline reference acquisition: about 2.4 days on 8$\times$ H100s for FLUX and 2.3 days on 8$\times$ RTX~4090s for Wan2.1-1.3B; gate optimization (about one minute) and deployment overhead are negligible by comparison. The search runs once per backbone, and the single resulting gate then serves every evaluated budget, prompt, and step count without further search or training. Furthermore, the offline search time can be adjusted to the available compute: cutting it to about half a day still yields acceptable quality (\cref{tab:labelcount}). All numbers in the paper use the full search.

\section{Conclusion}
\looseness=-1
\label{sec:conclusion}

\looseness=-1
Cache scheduling in diffusion models is a finite-horizon, budget-constrained allocation problem that demands global budget coordination and local trajectory awareness at once. Static schedules provide the former but run open-loop; online threshold rules provide the latter but leave the remaining budget and horizon out of the per-step rule. \bag{} combines the two by distilling offline, full-horizon searched reference schedules into a lightweight budget- and trajectory-conditioned online gate. This design makes computation predictable while preserving prompt-specific adaptation at inference. At matched computation, on held-out prompts, one gate per backbone outperforms strong static and online baselines across all evaluated budgets on image and video DiTs, remains effective under tested shifts of sampler step count, seed, resolution, and guidance, and spends exactly the requested budget by construction.

{
    \small
    \bibliographystyle{ieeenat_fullname}
    \bibliography{references}

\begin{thebibliography}{58}
\providecommand{\natexlab}[1]{#1}
\providecommand{\url}[1]{\texttt{#1}}
\expandafter\ifx\csname urlstyle\endcsname\relax
  \providecommand{\doi}[1]{doi: #1}\else
  \providecommand{\doi}{doi: \begingroup \urlstyle{rm}\Url}\fi

\bibitem[Aarts and Lenstra(2018)]{HC_opt}
Emile Aarts and Jan~Karel Lenstra.
\newblock \emph{Local search in combinatorial optimization}.
\newblock Princeton University Press, 2018.

\bibitem[Aliev et~al.(2026)Aliev, Neudachina, Bykov, Oganov, Struminsky,
  Alanov, and Rakitin]{aliev2026recache}
Mishan Aliev, Eva Neudachina, Ilya Bykov, Aleksandr Oganov, Kirill Struminsky,
  Aibek Alanov, and Denis Rakitin.
\newblock {ReCache}: Learning budget-aware caching schedules for diffusion
  models via {REINFORCE}.
\newblock \emph{arXiv preprint arXiv:2606.06060}, 2026.

\bibitem[Bu et~al.(2025)Bu, Ling, Zhou, Wang, Zang, Lin, and Wang]{dicache}
Jiazi Bu, Pengyang Ling, Yujie Zhou, Yibin Wang, Yuhang Zang, Dahua Lin, and
  Jiaqi Wang.
\newblock {DiCache}: Let diffusion model determine its own cache.
\newblock \emph{arXiv preprint arXiv:2508.17356}, 2025.

\bibitem[Cai et~al.(2025)Cai, Cao, Du, Gao, Hao, Hoi, Hou, Huang, Jiang, Jiang,
  et~al.]{zimage}
Huanqia Cai, Sihan Cao, Ruoyi Du, Peng Gao, Aiming Hao, Steven Hoi, Zhaohui
  Hou, Shijie Huang, Dengyang Jiang, Yuming Jiang, et~al.
\newblock {Z-Image}: An efficient image generation foundation model with
  single-stream diffusion transformer.
\newblock \emph{arXiv preprint arXiv:2511.22699}, 2025.

\bibitem[Chen et~al.(2025)Chen, Meng, Tang, Ma, Jiang, Wang, Wang, and
  Zhu]{qdit}
Lei Chen, Yuan Meng, Chen Tang, Xinzhu Ma, Jingyan Jiang, Xin Wang, Zhi Wang,
  and Wenwu Zhu.
\newblock {Q-DiT}: Accurate post-training quantization for diffusion
  transformers.
\newblock In \emph{Proceedings of the Computer Vision and Pattern Recognition
  Conference}, pages 28306--28315, 2025.

\bibitem[Chen et~al.(2024)Chen, Shen, Ye, Cao, Tu, Bouganis, Zhao, and
  Chen]{chen2024deltadit}
Pengtao Chen, Mingzhu Shen, Peng Ye, Jianjian Cao, Chongjun Tu, Christos-Savvas
  Bouganis, Yiren Zhao, and Tao Chen.
\newblock $\delta$-{DiT}: A training-free acceleration method tailored for
  diffusion transformers.
\newblock \emph{arXiv preprint arXiv:2406.01125}, 2024.

\bibitem[Chung et~al.(2026)Chung, Hyun, Lee, Han, Cha, Wee, Hong, and
  Heo]{chung2026seacache}
Jiwoo Chung, Sangeek Hyun, MinKyu Lee, Byeongju Han, Geonho Cha, Dongyoon Wee,
  Youngjun Hong, and Jae-Pil Heo.
\newblock {SeaCache}: Spectral-evolution-aware cache for accelerating diffusion
  models.
\newblock In \emph{Proceedings of the IEEE/CVF Conference on Computer Vision
  and Pattern Recognition (CVPR)}, pages 14283--14294, 2026.

\bibitem[Cui et~al.(2026)Cui, Wang, Xu, Chen, Zhang, Jiang, Jin, Liu, and
  Huang]{dpcache}
Bowen Cui, Yuanbin Wang, Huajiang Xu, Biaolong Chen, Aixi Zhang, Hao Jiang,
  Zhengzheng Jin, Xu Liu, and Pipei Huang.
\newblock Denoising as path planning: Training-free acceleration of diffusion
  models with {DPCache}.
\newblock \emph{arXiv preprint arXiv:2602.22654}, 2026.

\bibitem[Gao et~al.(2025)Gao, Chen, Shi, Tan, Liu, Zhao, Wang, and
  Lian]{lemica}
Huanlin Gao, Ping Chen, Fuyuan Shi, Chao Tan, Zhaoxiang Liu, Fang Zhao, Kai
  Wang, and Shiguo Lian.
\newblock {LeMiCa}: Lexicographic minimax path caching for efficient
  diffusion-based video generation.
\newblock \emph{arXiv preprint arXiv:2511.00090}, 2025.

\bibitem[Gao et~al.(2026)Gao, Chen, Shi, Wu, YanTao, Hui, You, Lu, Tan, Zhao,
  et~al.]{meancache}
Huanlin Gao, Ping Chen, Fuyuan Shi, Ruijia Wu, Li YanTao, Qiang Hui, Yuren You,
  Ting Lu, Chao Tan, Shaoan Zhao, et~al.
\newblock {MeanCache}: From instantaneous to average velocity for accelerating
  flow matching inference.
\newblock \emph{arXiv preprint arXiv:2601.19961}, 2026.

\bibitem[Ghosh et~al.(2023)Ghosh, Hajishirzi, and Schmidt]{ghosh2023geneval}
Dhruba Ghosh, Hannaneh Hajishirzi, and Ludwig Schmidt.
\newblock {GenEval}: An object-focused framework for evaluating text-to-image
  alignment.
\newblock \emph{Advances in Neural Information Processing Systems},
  36:\penalty0 52132--52152, 2023.

\bibitem[Haghighi and Alahi(2026)]{sencache}
Yasaman Haghighi and Alexandre Alahi.
\newblock {SenCache}: Accelerating diffusion model inference via
  sensitivity-aware caching.
\newblock \emph{arXiv preprint arXiv:2602.24208}, 2026.

\bibitem[Ho et~al.(2020)Ho, Jain, and Abbeel]{ho2020ddpm}
Jonathan Ho, Ajay Jain, and Pieter Abbeel.
\newblock Denoising diffusion probabilistic models.
\newblock \emph{Advances in neural information processing systems},
  33:\penalty0 6840--6851, 2020.

\bibitem[Huang et~al.(2024)Huang, He, Yu, Zhang, Si, Jiang, Zhang, Wu, Jin,
  Chanpaisit, et~al.]{huang2024vbench}
Ziqi Huang, Yinan He, Jiashuo Yu, Fan Zhang, Chenyang Si, Yuming Jiang, Yuanhan
  Zhang, Tianxing Wu, Qingyang Jin, Nattapol Chanpaisit, et~al.
\newblock {VBench}: Comprehensive benchmark suite for video generative models.
\newblock In \emph{Proceedings of the IEEE/CVF Conference on Computer Vision
  and Pattern Recognition}, pages 21807--21818, 2024.

\bibitem[Huang et~al.(2025)Huang, Yang, and Ren]{prompttea}
Zishen Huang, Chunyu Yang, and Mengyuan Ren.
\newblock {PromptTea}: Let prompts tell {TeaCache} the optimal threshold.
\newblock \emph{arXiv preprint arXiv:2507.06739}, 2025.

\bibitem[Kahatapitiya et~al.(2025)Kahatapitiya, Liu, He, Liu, Jia, Zhang, Ryoo,
  and Xie]{kahatapitiya2024adacache}
Kumara Kahatapitiya, Haozhe Liu, Sen He, Ding Liu, Menglin Jia, Chenyang Zhang,
  Michael~S Ryoo, and Tian Xie.
\newblock Adaptive caching for faster video generation with diffusion
  transformers.
\newblock In \emph{Proceedings of the IEEE/CVF International Conference on
  Computer Vision}, pages 15240--15252, 2025.

\bibitem[Kirkpatrick et~al.(1983)Kirkpatrick, Gelatt~Jr, and Vecchi]{SA_opt}
Scott Kirkpatrick, C~Daniel Gelatt~Jr, and Mario~P Vecchi.
\newblock Optimization by simulated annealing.
\newblock \emph{science}, 220\penalty0 (4598):\penalty0 671--680, 1983.

\bibitem[Kong et~al.(2024)Kong, Tian, Zhang, Min, Dai, Zhou, Xiong, Li, Wu,
  Zhang, et~al.]{hunyuanvideo}
Weijie Kong, Qi Tian, Zijian Zhang, Rox Min, Zuozhuo Dai, Jin Zhou, Jiangfeng
  Xiong, Xin Li, Bo Wu, Jianwei Zhang, et~al.
\newblock {HunyuanVideo}: A systematic framework for large video generative
  models.
\newblock \emph{arXiv preprint arXiv:2412.03603}, 2024.

\bibitem[Labs(2024)]{flux2024}
Black~Forest Labs.
\newblock {FLUX}.
\newblock \url{https://github.com/black-forest-labs/flux}, 2024.

\bibitem[Lei et~al.(2026)Lei, Zhao, Yuan, and Zhang]{lei2026budcache}
Mingkun Lei, Tong Zhao, Liangyu Yuan, and Chi Zhang.
\newblock Budget-constrained step-level diffusion caching.
\newblock In \emph{Forty-third International Conference on Machine Learning},
  2026.

\bibitem[Li et~al.(2023)Li, Liu, Lian, Yang, Dong, Kang, Zhang, and
  Keutzer]{qdiff}
Xiuyu Li, Yijiang Liu, Long Lian, Huanrui Yang, Zhen Dong, Daniel Kang,
  Shanghang Zhang, and Kurt Keutzer.
\newblock {Q-Diffusion}: Quantizing diffusion models.
\newblock In \emph{Proceedings of the IEEE/CVF International Conference on
  Computer Vision}, pages 17535--17545, 2023.

\bibitem[Lipman et~al.(2023)Lipman, Chen, Ben-Hamu, Nickel, and
  Le]{lipman2022flow}
Yaron Lipman, Ricky~TQ Chen, Heli Ben-Hamu, Maximilian Nickel, and Matthew Le.
\newblock Flow matching for generative modeling.
\newblock In \emph{The eleventh international conference on learning
  representations}, 2023.

\bibitem[Liu et~al.(2025{\natexlab{a}})Liu, Zhang, Wang, Wei, Qiu, Zhao, Zhang,
  Ye, and Wan]{liu2025teacache}
Feng Liu, Shiwei Zhang, Xiaofeng Wang, Yujie Wei, Haonan Qiu, Yuzhong Zhao,
  Yingya Zhang, Qixiang Ye, and Fang Wan.
\newblock Timestep embedding tells: It's time to cache for video diffusion
  model.
\newblock In \emph{Proceedings of the Computer Vision and Pattern Recognition
  Conference}, pages 7353--7363, 2025{\natexlab{a}}.

\bibitem[Liu et~al.(2025{\natexlab{b}})Liu, Zou, Lyu, Chen, and
  Zhang]{liu2025taylorseer}
Jiacheng Liu, Chang Zou, Yuanhuiyi Lyu, Junjie Chen, and Linfeng Zhang.
\newblock From reusing to forecasting: Accelerating diffusion models with
  {TaylorSeers}.
\newblock In \emph{Proceedings of the IEEE/CVF International Conference on
  Computer Vision}, pages 15853--15863, 2025{\natexlab{b}}.

\bibitem[Liu et~al.(2025{\natexlab{c}})Liu, Li, and Gu]{cachequant}
Xuewen Liu, Zhikai Li, and Qingyi Gu.
\newblock {CacheQuant}: Comprehensively accelerated diffusion models.
\newblock In \emph{Proceedings of the Computer Vision and Pattern Recognition
  Conference}, pages 23269--23280, 2025{\natexlab{c}}.

\bibitem[Lu et~al.(2022)Lu, Zhou, Bao, Chen, Li, and Zhu]{lu2022dpmsolver}
Cheng Lu, Yuhao Zhou, Fan Bao, Jianfei Chen, Chongxuan Li, and Jun Zhu.
\newblock {DPM-Solver}: A fast {ODE} solver for diffusion probabilistic model
  sampling in around 10 steps.
\newblock \emph{Advances in neural information processing systems},
  35:\penalty0 5775--5787, 2022.

\bibitem[Lyu et~al.(2025)Lyu, Si, Song, Yang, Qiao, Liu, and
  Wong]{lv2024fastercache}
Zhengyao Lyu, Chenyang Si, Junhao Song, Zhenyu Yang, Yu Qiao, Ziwei Liu, and
  Kwan-Yee~K Wong.
\newblock {FasterCache}: Training-free video diffusion model acceleration with
  high quality.
\newblock In \emph{International Conference on Learning Representations}, pages
  33132--33156, 2025.

\bibitem[Ma et~al.(2024{\natexlab{a}})Ma, Fang, Bi~Mi, and Wang]{ma2024l2c}
Xinyin Ma, Gongfan Fang, Michael Bi~Mi, and Xinchao Wang.
\newblock {Learning-to-Cache}: Accelerating diffusion transformer via layer
  caching.
\newblock \emph{Advances in Neural Information Processing Systems},
  37:\penalty0 133282--133304, 2024{\natexlab{a}}.

\bibitem[Ma et~al.(2024{\natexlab{b}})Ma, Fang, and Wang]{ma2024deepcache}
Xinyin Ma, Gongfan Fang, and Xinchao Wang.
\newblock {DeepCache}: Accelerating diffusion models for free.
\newblock In \emph{Proceedings of the IEEE/CVF conference on computer vision
  and pattern recognition}, pages 15762--15772, 2024{\natexlab{b}}.

\bibitem[Ma et~al.(2025)Ma, Wei, Wang, Zhang, and Tian]{ma2025magcache}
Zehong Ma, Longhui Wei, Feng Wang, Shiliang Zhang, and Qi Tian.
\newblock {MagCache}: Fast video generation with magnitude-aware cache.
\newblock \emph{Advances in Neural Information Processing Systems},
  38:\penalty0 34348--34380, 2025.

\bibitem[Peebles and Xie(2023)]{peebles2023dit}
William Peebles and Saining Xie.
\newblock Scalable diffusion models with transformers.
\newblock In \emph{Proceedings of the IEEE/CVF international conference on
  computer vision}, pages 4195--4205, 2023.

\bibitem[Peng et~al.(2025)Peng, Yan, Liu, Ma, Chen, Wang, Wu, Liu, and
  Lin]{ertacache}
Xurui Peng, Chenqian Yan, Hong Liu, Rui Ma, Fangmin Chen, Xing Wang, Zhihua Wu,
  Songwei Liu, and Mingbao Lin.
\newblock {ERTACache}: Error rectification and timesteps adjustment for
  efficient diffusion.
\newblock \emph{arXiv preprint arXiv:2508.21091}, 2025.

\bibitem[Radford et~al.(2021)Radford, Kim, Hallacy, Ramesh, Goh, Agarwal,
  Sastry, Askell, Mishkin, Clark, et~al.]{clip}
Alec Radford, Jong~Wook Kim, Chris Hallacy, Aditya Ramesh, Gabriel Goh,
  Sandhini Agarwal, Girish Sastry, Amanda Askell, Pamela Mishkin, Jack Clark,
  et~al.
\newblock Learning transferable visual models from natural language
  supervision.
\newblock In \emph{International conference on machine learning}, pages
  8748--8763. PmLR, 2021.

\bibitem[Ren et~al.(2024)Ren, Xia, Lu, Zhang, Wu, Xie, Wang, and Xiao]{hypersd}
Yuxi Ren, Xin Xia, Yanzuo Lu, Jiacheng Zhang, Jie Wu, Pan Xie, Xing Wang, and
  Xuefeng Xiao.
\newblock {Hyper-SD}: Trajectory segmented consistency model for efficient
  image synthesis.
\newblock \emph{Advances in neural information processing systems},
  37:\penalty0 117340--117362, 2024.

\bibitem[Ross et~al.(2011)Ross, Gordon, and Bagnell]{ross2011dagger}
St{\'e}phane Ross, Geoffrey Gordon, and Drew Bagnell.
\newblock A reduction of imitation learning and structured prediction to
  no-regret online learning.
\newblock In \emph{Proceedings of the fourteenth international conference on
  artificial intelligence and statistics}, pages 627--635. JMLR Workshop and
  Conference Proceedings, 2011.

\bibitem[Saharia et~al.(2022)Saharia, Chan, Saxena, Li, Whang, Denton,
  Ghasemipour, Gontijo~Lopes, Karagol~Ayan, Salimans,
  et~al.]{saharia2022imagen}
Chitwan Saharia, William Chan, Saurabh Saxena, Lala Li, Jay Whang, Emily~L
  Denton, Kamyar Ghasemipour, Raphael Gontijo~Lopes, Burcu Karagol~Ayan, Tim
  Salimans, et~al.
\newblock Photorealistic text-to-image diffusion models with deep language
  understanding.
\newblock \emph{Advances in neural information processing systems},
  35:\penalty0 36479--36494, 2022.

\bibitem[Salimans and Ho(2022)]{salimans2022progressive}
Tim Salimans and Jonathan Ho.
\newblock Progressive distillation for fast sampling of diffusion models.
\newblock \emph{arXiv preprint arXiv:2202.00512}, 2022.

\bibitem[Selvaraju et~al.(2024)Selvaraju, Ding, Chen, Zharkov, and
  Liang]{selvaraju2024fora}
Pratheba Selvaraju, Tianyu Ding, Tianyi Chen, Ilya Zharkov, and Luming Liang.
\newblock {FORA}: Fast-forward caching in diffusion transformer acceleration.
\newblock \emph{arXiv preprint arXiv:2407.01425}, 2024.

\bibitem[Song et~al.(2023)Song, Dhariwal, Chen, and
  Sutskever]{song2023consistency}
Yang Song, Prafulla Dhariwal, Mark Chen, and Ilya Sutskever.
\newblock Consistency models.
\newblock In \emph{Proceedings of the 40th International Conference on Machine
  Learning}, pages 32211--32252, 2023.

\bibitem[Team et~al.(2025)Team, Cai, Huang, Kang, Li, Liang, Ma, Ren, Wei, Xie,
  et~al.]{longcatvideo}
Meituan~LongCat Team, Xunliang Cai, Qilong Huang, Zhuoliang Kang, Hongyu Li,
  Shijun Liang, Liya Ma, Siyu Ren, Xiaoming Wei, Rixu Xie, et~al.
\newblock {LongCat-Video} technical report.
\newblock \emph{arXiv preprint arXiv:2510.22200}, 2025.

\bibitem[{Team Wan} et~al.(2025){Team Wan}, Wang, Ai, Wen, Mao, Xie, Chen, Yu,
  Zhao, Yang, Zeng, et~al.]{wan2025}
{Team Wan}, Ang Wang, Baole Ai, Bin Wen, Chaojie Mao, Chen-Wei Xie, Di Chen,
  Feiwu Yu, Haiming Zhao, Jianxiao Yang, Jianyuan Zeng, et~al.
\newblock Wan: Open and advanced large-scale video generative models.
\newblock \emph{arXiv preprint arXiv:2503.20314}, 2025.

\bibitem[Wang et~al.(2025)Wang, Zhu, Li, Yuan, and Zhang]{wang2026adaptive}
Ruoyu Wang, Beier Zhu, Junzhi Li, Liangyu Yuan, and Chi Zhang.
\newblock Adaptive stochastic coefficients for accelerating diffusion sampling.
\newblock \emph{Advances in Neural Information Processing Systems},
  38:\penalty0 21985--22016, 2025.

\bibitem[Wang et~al.(2004)Wang, Bovik, Sheikh, and Simoncelli]{ssim}
Zhou Wang, Alan~C Bovik, Hamid~R Sheikh, and Eero~P Simoncelli.
\newblock Image quality assessment: from error visibility to structural
  similarity.
\newblock \emph{IEEE transactions on image processing}, 13\penalty0
  (4):\penalty0 600--612, 2004.

\bibitem[Wimbauer et~al.(2024)Wimbauer, Wu, Schoenfeld, Dai, Hou, He,
  Sanakoyeu, Zhang, Tsai, Kohler, et~al.]{cachemeifyoucan}
Felix Wimbauer, Bichen Wu, Edgar Schoenfeld, Xiaoliang Dai, Ji Hou, Zijian He,
  Artsiom Sanakoyeu, Peizhao Zhang, Sam Tsai, Jonas Kohler, et~al.
\newblock Cache me if you can: Accelerating diffusion models through block
  caching.
\newblock In \emph{Proceedings of the IEEE/CVF Conference on Computer Vision
  and Pattern Recognition}, pages 6211--6220, 2024.

\bibitem[Wu et~al.(2025)Wu, Li, Zhou, Lin, Gao, Yan, Yin, Bai, Xu, Chen,
  et~al.]{qwenimage}
Chenfei Wu, Jiahao Li, Jingren Zhou, Junyang Lin, Kaiyuan Gao, Kun Yan,
  Sheng-ming Yin, Shuai Bai, Xiao Xu, Yilei Chen, et~al.
\newblock {Qwen-Image} technical report.
\newblock \emph{arXiv preprint arXiv:2508.02324}, 2025.

\bibitem[Wu et~al.(2023)Wu, Hao, Sun, Chen, Zhu, Zhao, and Li]{hpsv2}
Xiaoshi Wu, Yiming Hao, Keqiang Sun, Yixiong Chen, Feng Zhu, Rui Zhao, and
  Hongsheng Li.
\newblock Human preference score v2: A solid benchmark for evaluating human
  preferences of text-to-image synthesis.
\newblock \emph{arXiv preprint arXiv:2306.09341}, 2023.

\bibitem[Xu et~al.(2023)Xu, Liu, Wu, Tong, Li, Ding, Tang, and
  Dong]{xu2023imagereward}
Jiazheng Xu, Xiao Liu, Yuchen Wu, Yuxuan Tong, Qinkai Li, Ming Ding, Jie Tang,
  and Yuxiao Dong.
\newblock {ImageReward}: Learning and evaluating human preferences for
  text-to-image generation.
\newblock \emph{Advances in Neural Information Processing Systems},
  36:\penalty0 15903--15935, 2023.

\bibitem[Yin et~al.(2024{\natexlab{a}})Yin, Gharbi, Park, Zhang, Shechtman,
  Durand, and Freeman]{dmd2}
Tianwei Yin, Micha{\"e}l Gharbi, Taesung Park, Richard Zhang, Eli Shechtman,
  Fredo Durand, and Bill Freeman.
\newblock Improved distribution matching distillation for fast image synthesis.
\newblock \emph{Advances in neural information processing systems},
  37:\penalty0 47455--47487, 2024{\natexlab{a}}.

\bibitem[Yin et~al.(2024{\natexlab{b}})Yin, Gharbi, Zhang, Shechtman, Durand,
  Freeman, and Park]{dmd}
Tianwei Yin, Micha{\"e}l Gharbi, Richard Zhang, Eli Shechtman, Fredo Durand,
  William~T Freeman, and Taesung Park.
\newblock One-step diffusion with distribution matching distillation.
\newblock In \emph{Proceedings of the IEEE/CVF conference on computer vision
  and pattern recognition}, pages 6613--6623, 2024{\natexlab{b}}.

\bibitem[Yuan et~al.(2026)Yuan, Wang, Zhao, Fu, Lei, Zhu, and
  Zhang]{Yuan_2026_CVPR}
Liangyu Yuan, Ruoyu Wang, Tong Zhao, Dingwen Fu, Mingkun Lei, Beier Zhu, and
  Chi Zhang.
\newblock Few-step diffusion sampling through instance-aware discretizations.
\newblock In \emph{Proceedings of the IEEE/CVF Conference on Computer Vision
  and Pattern Recognition (CVPR)}, pages 35882--35892, 2026.

\bibitem[Zhang et~al.(2025)Zhang, Gao, Shao, and Wu]{blockdance}
Hui Zhang, Tingwei Gao, Jie Shao, and Zuxuan Wu.
\newblock {BlockDance}: Reuse structurally similar spatio-temporal features to
  accelerate diffusion transformers.
\newblock In \emph{Proceedings of the Computer Vision and Pattern Recognition
  Conference}, pages 12891--12900, 2025.

\bibitem[Zhang et~al.(2018)Zhang, Isola, Efros, Shechtman, and
  Wang]{zhang2018lpips}
Richard Zhang, Phillip Isola, Alexei~A Efros, Eli Shechtman, and Oliver Wang.
\newblock The unreasonable effectiveness of deep features as a perceptual
  metric.
\newblock In \emph{Proceedings of the IEEE conference on computer vision and
  pattern recognition}, pages 586--595, 2018.

\bibitem[Zhao et~al.(2026)Zhao, Lei, Yuan, Yang, Song, Wang, Zhu, and
  Zhang]{zhao2026dyweight}
Tong Zhao, Mingkun Lei, Liangyu Yuan, Yanming Yang, Chenxi Song, Yang Wang,
  Beier Zhu, and Chi Zhang.
\newblock {DyWeight}: Dynamic gradient weighting for few-step diffusion
  sampling.
\newblock \emph{arXiv preprint arXiv:2603.11607}, 2026.

\bibitem[Zhao et~al.(2023)Zhao, Bai, Rao, Zhou, and Lu]{zhao2023unipc}
Wenliang Zhao, Lujia Bai, Yongming Rao, Jie Zhou, and Jiwen Lu.
\newblock {UniPC}: A unified predictor-corrector framework for fast sampling of
  diffusion models.
\newblock \emph{Advances in Neural Information Processing Systems},
  36:\penalty0 49842--49869, 2023.

\bibitem[Zhao et~al.(2025{\natexlab{a}})Zhao, Han, Tang, Wang, Song, Huang,
  Wang, and You]{dydit}
Wangbo Zhao, Yizeng Han, Jiasheng Tang, Kai Wang, Yibing Song, Gao Huang, Fan
  Wang, and Yang You.
\newblock Dynamic diffusion transformer.
\newblock In \emph{International Conference on Learning Representations}, pages
  65520--65552, 2025{\natexlab{a}}.

\bibitem[Zhao et~al.(2025{\natexlab{b}})Zhao, Jin, Wang, and You]{zhao2024pab}
Xuanlei Zhao, Xiaolong Jin, Kai Wang, and Yang You.
\newblock Real-time video generation with pyramid attention broadcast.
\newblock In \emph{International Conference on Learning Representations}, pages
  3296--3319, 2025{\natexlab{b}}.

\bibitem[Zhu et~al.(2025)Zhu, Wang, Zhao, Zhang, and Zhang]{zhu2025distilling}
Beier Zhu, Ruoyu Wang, Tong Zhao, Hanwang Zhang, and Chi Zhang.
\newblock Distilling parallel gradients for fast {ODE} solvers of diffusion
  models.
\newblock In \emph{2025 IEEE/CVF International Conference on Computer Vision
  (ICCV)}, pages 19557--19566. IEEE, 2025.

\bibitem[Zou et~al.(2025)Zou, Liu, Liu, Huang, and Zhang]{zou2024toca}
Chang Zou, Xuyang Liu, Ting Liu, Siteng Huang, and Linfeng Zhang.
\newblock Accelerating diffusion transformers with token-wise feature caching.
\newblock In \emph{The Thirteenth International Conference on Learning
  Representations}, 2025.

\end{thebibliography}
}

\twocolumn[%
  \begin{center}
    {\LARGE\bfseries \papertitle\par}
    \vskip 0.12in
    {\Large\bfseries Supplementary Material\par}
    \vskip 0.18in
  \end{center}%
]
\appendix
\crefalias{section}{appendix}
\crefalias{subsection}{appendix}
\setcounter{secnumdepth}{2}
\setcounter{table}{0}
\setcounter{figure}{0}
\renewcommand{\thetable}{A\arabic{table}}
\renewcommand{\thefigure}{A\arabic{figure}}

\section*{Contents}
\noindent
\begin{tabular}{@{}p{0.5cm}l@{}}
\textbf{\ref{app:impl}} & Implementation Details \\
\textbf{\ref{app:baselines}} & Baseline Details \\
\textbf{\ref{app:tables}} & Additional Quantitative Results \\
\textbf{\ref{app:qual}} & Additional Qualitative Results \\
\textbf{\ref{app:discussion}} & Limitations \\
\end{tabular}

\section{Implementation Details}
\label{app:impl}

\subsection{Offline Search Protocol}
\label{app:search}

\paragraph{Objective and evaluator.} For each (prompt, budget) cell, we minimize LPIPS~\citep{zhang2018lpips} to the same-seed full-compute output over binary masks $\mathbf{m}$ with $\|\mathbf{m}\|_1{=}B$ and $m_0{=}1$. Under classifier-free guidance, the two branches share one mask: a step is either computed for both or reused for both. The reuse mechanism is the standard residual path; the only change is that the decision comes from the mask instead of a threshold test. Each prompt keeps one fixed seed, shared by all methods and the full-compute reference ($42$ on FLUX~\citep{flux2024}, $42$ plus the prompt index on Wan~\citep{wan2025}), and every mask rollout runs on the same GPU as its full-compute target, since outputs are reproducible only within one machine.

\paragraph{Optimizer.} The optimizer is a simple multi-start local search over masks, run for each (prompt, budget) cell. The starting points include: the prompt's own SeaCache~\citep{chung2026seacache} schedule at the nearest threshold, adjusted to exactly $B$ computed steps; masks that spend most of the budget early; a uniform mask; and, at higher budgets, the reference from the next lower budget, with the extra computed steps inserted at reused steps (evenly spaced on FLUX, at random on Wan). A move shifts or relocates one computed step, and every evaluated mask's score is cached, because the output is deterministic given the mask. On FLUX~\citep{flux2024}, we run a short simulated-annealing~\citep{SA_opt} phase from the three best starts, then greedy refinement; a cell takes about 450 rollouts on average. On Wan, where a rollout generates a full video, we run greedy refinement alone, about 100 rollouts per cell, and add an annealing burst when refinement brings little improvement. The returned masks $\hat{\mathbf{m}}$ are the best found, not proven optima. Qwen-Image-2512~\citep{qwenimage} follows the same offline search protocol and budget grid as FLUX, with the same calibration prompts.

\subsection{Gate Training}
\label{app:gate}

\looseness=-1
\cref{alg:train} gives the training pipeline end to end: the two loops collect the offline references and turn each reference rollout into per-step examples, and the last two lines fit the gate that \cref{alg:gate} then deploys. Only the search inside the inner loop is expensive, and it runs once per backbone, never at deployment. Both gates use training-prompt z-score statistics and class-balanced BCE, weighting positives by the training-set negative-to-positive ratio. The six inputs and the single output are the same on both backbones; only the hidden widths differ. On FLUX~\citep{flux2024} the gate is one narrow MLP over all six features ($6{\to}8{\to}8$ followed by $8{\to}32{\to}1$; $449$ parameters). On Wan it has one branch per signal group (two $3{\to}8{\to}8$ branches fused by $16{\to}32{\to}1$; $785$ parameters). The Qwen-Image gate reuses the FLUX architecture. Both gates are trained with AdamW (learning rate $10^{-3}$, batch size 8192); weight decay is $10^{-4}$ on FLUX and zero on Wan. The FLUX gate uses a prompt-level train/validation split, trains for up to 300 epochs, and deploys the checkpoint with the best validation AUC. The Wan labeled set is too small to hold out a validation split, so its gate trains on all 50 prompts for a fixed 300 epochs, chosen where the training AUC plateaus (at batch size 8192 over the ${\sim}$10K Wan examples this is only ${\sim}$600 optimizer steps). Both gates train in about one minute on one RTX~4090; one checkpoint per backbone produces every \bag{} number.

\begin{table*}[t]
\centering
\footnotesize
\tabcolsep=14pt
\begin{tabular}{lllll}
\toprule
Paradigm & Deployed object & Online state & Training signal & New budgets \\
\midrule
Static schedules & binary mask & no & optional, offline & new search \\
Threshold rules & hand-crafted rule & yes & none & threshold sweep \\
\bag{} (ours) & learned gate & yes & searched references & budget input \\
\bottomrule
\end{tabular}
\caption{\textbf{Paradigm summary: what is deployed for cache scheduling.} Static schedules replay one mask for every prompt; threshold rules decide online through a hand-crafted readout; \bag{} deploys a learned gate that decides online from budget and trajectory state, supervised by searched reference decisions, and takes the budget as a direct input.}
\label{tab:paradigm}
\end{table*}

\paragraph{Off-policy distribution shift.} Labels are recorded along the reference rollouts, but at deployment the gate follows its own past decisions, so it sees different states than the ones it was trained on. All reported results are measured under deployment and therefore already include the cost of this shift. On-policy correction in the style of DAgger~\citep{ross2011dagger} could reduce it, but every round needs fresh rollouts and labels, so we leave it to future work.

\subsection{Prompt Isolation}
\label{app:isolation}

On FLUX, the training prompts come from GenEval and the evaluation prompts from DrawBench, two disjoint datasets. On Wan, both sets come from the same VBench prompt pool, so we check them for three forms of overlap: identical prompts, substring pairs (\eg{} ``a stop sign'' vs.\ ``In a still frame, a stop sign''), and prompts that share a base sentence but differ in style or camera suffixes (\eg{} ``\ldots{}cafe in Paris'' vs.\ ``\ldots{}cafe in Paris, watercolor''). Training prompts that overlap with the evaluation set in any of these forms are replaced, so the final 50 training prompts have no exact-prompt, substring, or base-sentence overlap with the 100 evaluation prompts.

\begin{table}[!t]
\centering
\footnotesize
\tabcolsep=6pt
\begin{tabular}{lccc}
\toprule
 & \multicolumn{3}{c}{\bag{} improvement over SeaCache} \\
\cmidrule(lr){2-4}
 & PSNR gain (dB) & SSIM gain & LPIPS reduction \\
\midrule
\multicolumn{4}{@{}l}{\emph{FLUX.1-dev (200 prompts)}} \\
\midrule
$B{=}9$  & $+1.61^{***}$ & $+0.011^{***}$ & $0.014^{*}$ \\
$B{=}13$ & $+2.61^{***}$ & $+0.028^{***}$ & $0.036^{***}$ \\
$B{=}20$ & $+1.37^{***}$ & $+0.004^{\textrm{ns}}$ & $0.006^{\textrm{ns}}$ \\
\midrule
\multicolumn{4}{@{}l}{\emph{Wan2.1 (100 prompts)}} \\
\midrule
$B{=}15$ & $+1.97^{***}$ & $+0.043^{***}$ & $0.033^{***}$ \\
$B{=}19$ & $+1.61^{***}$ & $+0.026^{***}$ & $0.019^{***}$ \\
$B{=}24$ & $+0.66^{**}$  & $+0.004^{\textrm{ns}}$ & $0.001^{\textrm{ns}}$ \\
\bottomrule
\end{tabular}
\caption{\textbf{Paired gains of \bag{} over SeaCache on held-out prompts.} Each entry is the paired mean difference on the same prompts using the same or fewer NFEs, oriented so that higher is better: PSNR/SSIM gains are \bag{}$\,-\,$SeaCache, LPIPS reduction is SeaCache$\,-\,$\bag{}. Every entry favors \bag{}. Superscripts indicate $p$-value ranges from two-sided paired Wilcoxon tests: $^{***}p{<}10^{-3}$, $^{**}p{<}10^{-2}$, $^{*}p{<}5{\times}10^{-2}$, and $^{\textrm{ns}}$ not significant.}
\label{tab:sig}
\end{table}

\section{Baseline Details}
\label{app:baselines}

\paragraph{Operating points.} \looseness=-2 Per-tier operating points are chosen by sweeping each method's threshold on the test set to match the target NFE. On FLUX: TeaCache $\delta{=}1.0/0.7/0.38$; MagCache $\delta{=}0.8/0.32/0.06$ with official calibration and $K{=}7/5/5$ ($K{=}5$ cannot reach the tightest tier); SeaCache $\delta{=}1.0/0.6/0.3$. TaylorSeer uses official bf16 first-order code. Its fixed refresh intervals omit NFEs $13$ and $20$. We use the interval at or just above each target: intervals $7/4/2$ give NFE $9/14/26$ for the $9/13/20$ targets, and TaylorSeer never spends fewer evaluations than \bag{}. On Wan: TeaCache at its recommended 1.3B setting, $\delta{=}0.18/0.14/0.09$ (realized NFE 17/20/25); MagCache with official ratios, $K{=}5$, and retention $0.1$ at the tightest tier (the default $0.2$ cannot reach NFE 15); SeaCache $\delta{=}0.44/0.29/0.19$. On Qwen-Image: SeaCache $\delta{=}1.6/0.7$ for the $B{=}9/15$ tiers (realized NFE $9.0/15.0$).

\paragraph{BudCache under its official protocol.} For completeness we also ran BudCache exactly as proposed: its simulated annealing plus hill-climbing search with the final-latent-MSE objective on its own calibration prompt, broadcast to all test prompts. This gives PSNR/SSIM/LPIPS of $20.50/0.747/0.306$ at $B{=}9$, $20.98/0.789/0.235$ at $B{=}13$, and $26.02/0.887/0.110$ at $B{=}20$. It is behind \bag{} at every tier. It is also behind the variant in the main tables at $B{=}13$ and $B{=}20$, and ahead of it at $B{=}9$. The main tables use the variant with our search, so the gap to \bag{} comes from broadcasting one schedule, not from a weaker search.

\paragraph{Forecasting-based caching.} \looseness=-2 TaylorSeer changes reuse rather than scheduling: skipped steps extrapolate cached features with a Taylor expansion. It therefore complements \bag{}. At realized NFEs of $9$, $14$, and $26$, it trails SeaCache by $3.6$--$4.3$\,dB PSNR (\cref{tab:flux}): when refreshes are far apart, the extrapolation error grows. Forecast cost per skipped step ranges from ${\approx}6\%$ of the denoise time at the loosest FLUX tier to ${\approx}25\%$ at the tightest. We leave the combination of \bag{}'s scheduling with such reuse-side correction methods to future work.

\paragraph{Threshold-to-compute mapping.} How much compute a threshold yields depends on the method and the setting. On FLUX at $T{=}50$, TeaCache moves from 21 to 26 NFEs as $\delta$ changes from $0.38$ to $0.3$, while MagCache at fixed $K{=}5$ moves only from 12 to 13 NFEs as $\delta$ changes from $0.8$ to $0.32$; SeaCache's $\delta{=}0.6$ spends 26\% of the steps at $T{=}50$ but 40\% at $T{=}25$. Reaching a target NFE therefore requires a sweep per deployment setting (sampler, resolution, backbone), whereas \bag{} accepts the budget as a direct input.

\paragraph{Paradigm summary.} \cref{tab:paradigm} summarizes what each paradigm deploys and how it handles budgets.

\begin{table}[!t]
\centering
\footnotesize
\tabcolsep=3.2pt
\begin{tabular}{lcccc}
\toprule
Method ($\sim$2.4$\times$ tier) & NFE & ImageReward$\uparrow$ & HPSv2$\uparrow$ & CLIP$\uparrow$ \\
\midrule
FLUX.1-dev & 50 & 1.008 & 0.3024 & 31.27 \\
\midrule
TeaCache & 21.0 & 0.991 & \underline{0.3015} & 31.18 \\
MagCache & 20.0 & \underline{0.994} & 0.3009 & 31.20 \\
BudCache$^\dagger$ & 20.0 & 0.974 & 0.3011 & \underline{31.30} \\
SeaCache & 20.9 & 0.978 & 0.3008 & 31.23 \\
\rowcolor{lightCyan}\bag{} (ours) & 20.0 & 0.976 & 0.3002 & 31.26 \\
\rowcolor{lightCyan}\bag{}-R (ours) & 20.0 & \textbf{1.021} & \textbf{0.3021} & \textbf{31.42} \\
\bottomrule
\end{tabular}
\caption{\textbf{Preference and alignment metrics on FLUX.1-dev} ($\sim$2.4$\times$ tier). \bag{}-R (\emph{reward-supervised}) uses the same gate architecture and training recipe as \bag{}, differing only in the search objective used to produce its reference schedules: HPSv2 in place of LPIPS to the 50-step output. \textbf{Bold}: best; \underline{underline}: second; top row: full-compute reference. $^\dagger$BudCache evaluated under our search protocol.}
\label{tab:reward}
\end{table}

\section{Additional Quantitative Results}
\label{app:tables}

\begin{table*}[t]
\centering
\footnotesize
\tabcolsep=8pt
\begin{tabular}{l c c c c c c c c}
\toprule
Method & Subject$\uparrow$ & Background$\uparrow$ & Smoothness$\uparrow$ & Flickering$\uparrow$ & Dynamic$\uparrow$ & Aesthetic$\uparrow$ & Imaging$\uparrow$ & Avg$\uparrow$ \\
\midrule
Wan2.1 (50) & 0.9734 & 0.9742 & 0.9893 & 0.9825 & 0.3400 & 0.6133 & 0.7017 & 0.7963 \\
\midrule
\multicolumn{9}{l}{\emph{$B{=}15$}} \\
\dashedmidrule{9}
TeaCache & 0.9730 & 0.9726 & 0.9888 & 0.9831 & 0.3000 & 0.6065 & 0.6950 & 0.7884 \\
MagCache & 0.9719 & 0.9732 & 0.9887 & 0.9832 & 0.2800 & 0.6010 & 0.6881 & 0.7837 \\
BudCache$^\dagger$ & 0.9738 & 0.9721 & 0.9886 & 0.9828 & 0.3000 & 0.6041 & 0.6969 & 0.7883 \\
SeaCache & 0.9720 & 0.9719 & 0.9888 & 0.9828 & 0.2700 & 0.5991 & 0.6935 & 0.7826 \\
\rowcolor{lightCyan}\bag{} (ours) & 0.9743 & 0.9722 & 0.9887 & 0.9828 & 0.3100 & 0.6046 & 0.6962 & 0.7898 \\
\midrule
\multicolumn{9}{l}{\emph{$B{=}19$}} \\
\dashedmidrule{9}
TeaCache & 0.9725 & 0.9743 & 0.9889 & 0.9828 & 0.3200 & 0.6087 & 0.6936 & 0.7915 \\
MagCache & 0.9742 & 0.9745 & 0.9890 & 0.9830 & 0.3200 & 0.6070 & 0.6977 & 0.7922 \\
BudCache$^\dagger$ & 0.9736 & 0.9742 & 0.9887 & 0.9827 & 0.3400 & 0.6088 & 0.6984 & 0.7952 \\
SeaCache & 0.9749 & 0.9738 & 0.9889 & 0.9828 & 0.3100 & 0.6063 & 0.6987 & 0.7908 \\
\rowcolor{lightCyan}\bag{} (ours) & 0.9747 & 0.9731 & 0.9887 & 0.9827 & 0.3300 & 0.6091 & 0.6987 & 0.7938 \\
\midrule
\multicolumn{9}{l}{\emph{$B{=}24$}} \\
\dashedmidrule{9}
TeaCache & 0.9727 & 0.9744 & 0.9891 & 0.9828 & 0.3200 & 0.6123 & 0.6975 & 0.7927 \\
MagCache & 0.9741 & 0.9743 & 0.9892 & 0.9829 & 0.3200 & 0.6097 & 0.6999 & 0.7929 \\
BudCache$^\dagger$ & 0.9740 & 0.9739 & 0.9889 & 0.9827 & 0.3400 & 0.6135 & 0.6986 & 0.7960 \\
SeaCache & 0.9745 & 0.9735 & 0.9891 & 0.9827 & 0.3300 & 0.6090 & 0.7004 & 0.7942 \\
\rowcolor{lightCyan}\bag{} (ours) & 0.9745 & 0.9742 & 0.9888 & 0.9826 & 0.3300 & 0.6113 & 0.6993 & 0.7944 \\
\bottomrule
\end{tabular}
\caption{\textbf{Reference-free VBench dimensions on Wan2.1} at all three tiers (custom-input protocol; the nine remaining VBench dimensions are defined only on the official prompt suite). No per-cell best markers: at every tier every method sits within $0.003$ of the reference on the first four dimensions. Top row: full-compute reference. $^\dagger$BudCache under our search.}
\label{tab:vbench}
\end{table*}

\begin{table}[!t]
\centering
\footnotesize
\tabcolsep=5pt
\begin{tabular}{lccc}
\toprule
Variant & PSNR$\uparrow$ & SSIM$\uparrow$ & LPIPS$\downarrow$ \\
\midrule
\multicolumn{4}{l}{\textbf{$B{=}9$}} \\
\dashedmidrule{4}
SeaCache & 19.66 & 0.755 & 0.303 \\
\bag{} & \textbf{21.27} & \textbf{0.766} & \textbf{0.289} \\
w/o budget input & 18.80 & 0.596 & 0.572 \\
w/o trajectory input & 19.12 & 0.707 & 0.412 \\
\midrule
\multicolumn{4}{l}{\textbf{$B{=}13$}} \\
\dashedmidrule{4}
SeaCache & 21.85 & 0.821 & 0.203 \\
\bag{} & \textbf{24.46} & \textbf{0.849} & \textbf{0.168} \\
w/o budget input & 20.21 & 0.643 & 0.499 \\
w/o trajectory input & 20.49 & 0.759 & 0.299 \\
\midrule
\multicolumn{4}{l}{\textbf{$B{=}20$}} \\
\dashedmidrule{4}
SeaCache & 27.81 & 0.914 & 0.084 \\
\bag{} & \textbf{29.18} & \textbf{0.918} & \textbf{0.077} \\
w/o budget input & 22.18 & 0.751 & 0.381 \\
w/o trajectory input & 26.32 & 0.888 & 0.117 \\
\bottomrule
\end{tabular}
\caption{\textbf{Feature ablation at all budgets} (same protocol as \cref{tab:abl}). The collapse repeats at every budget; removing the budget input is the most destructive everywhere.}
\label{tab:abl-full}
\end{table}

\begin{table}[t]
\centering
\footnotesize
\tabcolsep=4pt
\begin{tabular}{llccc}
\toprule
Trajectory signal & Decision rule & PSNR$\uparrow$ & SSIM$\uparrow$ & LPIPS$\downarrow$ \\
\midrule
SeaCache signals & threshold & 21.85 & 0.821 & 0.203 \\
SeaCache signals & \bag{} gate & \textbf{24.68} & 0.841 & 0.179 \\
\bag{} features & \bag{} gate & 24.46 & \textbf{0.849} & \textbf{0.168} \\
\bottomrule
\end{tabular}
\caption{\textbf{Alternative trajectory descriptor} (FLUX, $B{=}13$). We compare three configurations: SeaCache's signals with its original threshold rule, the same signals with a retrained \bag{} gate, and \bag{}'s default features with the \bag{} gate. The budget state, architecture, labels, and training procedure are held fixed.}
\label{tab:traj}
\end{table}

\paragraph{Paired gains over SeaCache.} \looseness=-1 Across the main-table cells, \cref{tab:sig} reports mean paired gains over SeaCache on the same prompts and two-sided paired Wilcoxon tests. \bag{} uses the same or fewer NFEs, and higher is better in every column. PSNR gains are significant at every tier on both models. Perceptual gains are significant at the tight and mid tiers; at the loosest tier ($B{=}20$ on FLUX, $B{=}24$ on Wan) they are small and not significant because both methods are within $0.09$ LPIPS of the full-compute output.

\paragraph{Preference and alignment metrics.} \looseness=-2 \cref{tab:reward} scores the $\sim$2.4$\times$ FLUX tier with ImageReward~\citep{xu2023imagereward}, HPSv2~\citep{hpsv2}, and CLIP score~\citep{clip}. These metrics assess preference and text alignment rather than reconstruction, and may disagree: a high-scoring image can be far from the 50-step output. Caching itself targets reconstruction: it reuses outputs on the premise that the network's output changes little between steps, so the goal is to reproduce the full-compute output, and the main tables report exactly this. \bag{}'s reference schedules are searched to minimize LPIPS to the 50-step output. On the preference metrics, all methods stay within $0.005$ HPSv2 of one another and of the full-compute output. The optimization target, however, is set by the labels, not by the method. To show this, we repeated the pipeline with one change: the search maximizes HPSv2 instead of minimizing LPIPS (60 GenEval prompts $\times$ 3 budgets), and the schedules are distilled with the same recipe. The resulting gate, \bag{}-R in \cref{tab:reward}, leads all compared methods on all three metrics; its ImageReward and CLIP also exceed the 50-step model's own, and its HPSv2 falls $0.0003$ short of it. Reconstruction drops accordingly: LPIPS at $B{=}20$ rises from $0.077$ to $0.241$. This trade-off is expected: the highest-preference image is generally not the one closest to the 50-step output. The variants differ only in their search objective: reconstruction or preference.

\paragraph{Reference-free quality.} \cref{tab:vbench} scores all three Wan tiers on the seven reference-free VBench dimensions defined for custom inputs. At every tier and for every method, subject consistency, background consistency, motion smoothness, and flickering stay within $0.003$ of the full-compute reference. The main difference is dynamic degree at $B{=}15$, where \bag{} stays closest to the reference ($0.31$ against SeaCache's $0.27$; reference $0.34$) and has the highest average among the accelerated methods; at $B{=}19$ and $24$ all methods are within $0.03$ of the reference.

\paragraph{Feature ablation at all budgets.} \cref{tab:abl-full} carries the main-text ablation to $B{=}20$. Removing either half of the state hurts at every budget: without the budget state, PSNR falls $2.5$--$7.0$\,dB below the full gate; without the trajectory state, it falls $2.2$--$4.0$\,dB below. Both variants also fall below SeaCache at every budget.

\paragraph{Alternative trajectory descriptor.} \bag{}'s three trajectory features are one choice among several. \cref{tab:traj} replaces them with SeaCache's signals (a spectrally filtered feature distance and its running accumulator) and retrains the gate on those inputs, keeping the budget state, architecture, labels, and training procedure fixed. Reading the same signals with the gate instead of a threshold gains $2.8$\,dB PSNR, and the result comes within about $0.2$\,dB of \bag{}'s default features, ahead on PSNR and behind on SSIM and LPIPS. What matters is that a budget-conditioned gate reads the signals, not which particular signals it reads.

\begin{table}[t]
\centering
\footnotesize
\tabcolsep=2.6pt
\begin{tabular}[t]{lcccccc}
\toprule
 & \multicolumn{3}{c}{$B{=}9$} & \multicolumn{3}{c}{$B{=}13$} \\
\cmidrule(lr){2-4}\cmidrule(lr){5-7}
Prompts & PSNR$\uparrow$ & SSIM$\uparrow$ & LPIPS$\downarrow$ & PSNR$\uparrow$ & SSIM$\uparrow$ & LPIPS$\downarrow$ \\
\midrule
96 (full) & 21.27 & 0.766 & 0.2887 & 24.46 & 0.849 & 0.1675 \\
48 & 20.85 & 0.765 & 0.2884 & 24.29 & 0.848 & 0.1703 \\
24 & 20.38 & 0.749 & 0.3046 & 24.23 & 0.843 & 0.1766 \\
\bottomrule
\end{tabular}
\caption{\textbf{Label-count ablation} (FLUX.1-dev). Rows subsample the number of label prompts, each still labeled at all seven budgets; columns report deployed quality at two budgets. Quality degrades gracefully as the label set shrinks.}
\label{tab:labelcount}
\end{table}

\paragraph{Label-count ablation.} \looseness=-1 \cref{tab:labelcount} varies the FLUX label prompts, each labeled at all seven budgets, so the bank shrinks from $96{\times}7$ to $24{\times}7$ cells. Fewer prompts shorten the search proportionally but gradually reduce quality. Compared with the full bank, the PSNR costs at $B{=}9/13$ are $0.4/0.2$\,dB with 48 prompts and $0.9/0.2$\,dB with 24 prompts (about half a day on 8 GPUs). Thus label count can follow available resources: more labels improve the gate, whereas fewer labels reduce offline cost. All results use the full bank.

\begin{table}[!t]
\centering
\footnotesize
\tabcolsep=3.4pt
\begin{tabular}[t]{ccccccc}
\toprule
 & \multicolumn{3}{c}{Naive step reduction} & \multicolumn{3}{c}{\bag{} (one checkpoint)} \\
\cmidrule(lr){2-4}\cmidrule(lr){5-7}
NFE & PSNR$\uparrow$ & SSIM$\uparrow$ & LPIPS$\downarrow$ & PSNR$\uparrow$ & SSIM$\uparrow$ & LPIPS$\downarrow$ \\
\midrule
10 & 13.93 & 0.574 & 0.536 & \textbf{21.97} & \textbf{0.771} & \textbf{0.268} \\
12 & 14.44 & 0.597 & 0.498 & \textbf{23.16} & \textbf{0.807} & \textbf{0.216} \\
14 & 14.84 & 0.614 & 0.475 & \textbf{24.41} & \textbf{0.836} & \textbf{0.176} \\
16 & 15.21 & 0.631 & 0.446 & \textbf{25.51} & \textbf{0.860} & \textbf{0.142} \\
18 & 15.80 & 0.658 & 0.405 & \textbf{26.37} & \textbf{0.877} & \textbf{0.118} \\
20 & 16.00 & 0.663 & 0.402 & \textbf{27.68} & \textbf{0.894} & \textbf{0.098} \\
22 & 16.58 & 0.685 & 0.364 & \textbf{28.61} & \textbf{0.911} & \textbf{0.080} \\
24 & 16.98 & 0.697 & 0.350 & \textbf{29.57} & \textbf{0.924} & \textbf{0.066} \\
\bottomrule
\end{tabular}
\caption{\textbf{Budget sweep of one \bag{} checkpoint vs.\ naive step reduction} (50 unseen prompts, $B{=}$NFE on the 50-step grid). Step reduction converges to a different output and barely improves with more steps; the single \bag{} gate improves monotonically across the whole range.}
\label{tab:sweep}
\vspace{-1em}
\end{table}
\paragraph{Budget sweep against step reduction.} The simplest way to spend less compute is to run fewer sampler steps. \cref{tab:sweep} compares the single FLUX checkpoint with step reduction at $B{=}10,12,\ldots,24$ on 50 further unseen prompts. Both methods use the same seed per prompt and are scored against the corresponding 50-step output. Few-step sampling settles on a visibly different image and improves little with more steps. \bag{} stays close to the full-compute output and improves steadily as the budget grows, leading step reduction by $8$--$13$\,dB PSNR at every budget while using a single \bag{} gate throughout. \cref{fig:sweep,fig:sweep2} show eight of these prompts.

\paragraph{Capacity.} A gate more than an order of magnitude larger (${\approx}20$K parameters against $449$) is no better at $B{=}13$: it gains $0.09$\,dB PSNR but loses $0.006$ SSIM and $0.006$ LPIPS. Making the gate larger therefore does not help: the decision rule behind good schedules is compact.

\paragraph{The $\tau$ cutoff.} Since the NFE is pinned at $B$, changing $\tau$ only moves \emph{which} steps are computed; the cost stays the same. We sweep $\tau$ over $\{0.3,\ldots,0.7\}$ at FLUX $B{=}13$: PSNR falls as $\tau$ rises ($24.78 \to 23.45$), and LPIPS is lowest at the default ($0.188 \to 0.177 \to \mathbf{0.168} \to 0.171 \to 0.177$). A lower $\tau$ computes more of the budget early, which favors pixel fidelity; a higher $\tau$ spends it too late to help. All main results use $\tau{=}0.5$, fixed in advance; this sweep is a sensitivity check, and the default gives the best LPIPS of the values tested.

\section{Additional Qualitative Results}
\label{app:qual}

\begin{figure}[t]
\centering
\includegraphics[width=\columnwidth]{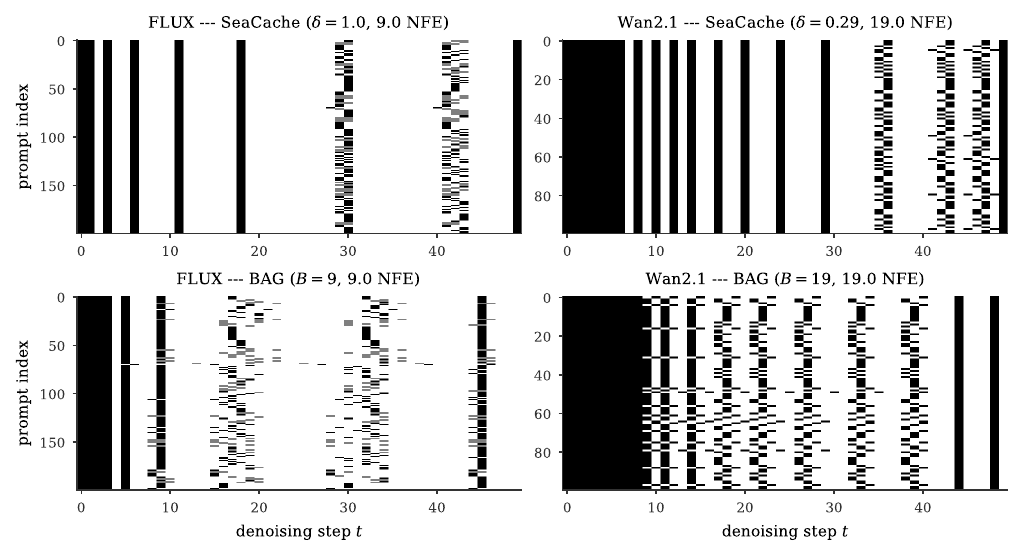}
\caption{\textbf{Realized compute placement at matched NFE} (left: FLUX; right: Wan2.1; black = computed). SeaCache follows the pattern induced by its accumulated-threshold rule, while \bag{} produces schedules from the learned budget- and trajectory-conditioned gate. Both methods use the same number of function evaluations.}
\label{fig:masks}
\end{figure}

\begin{figure*}[t]
\centering
\includegraphics[width=\textwidth]{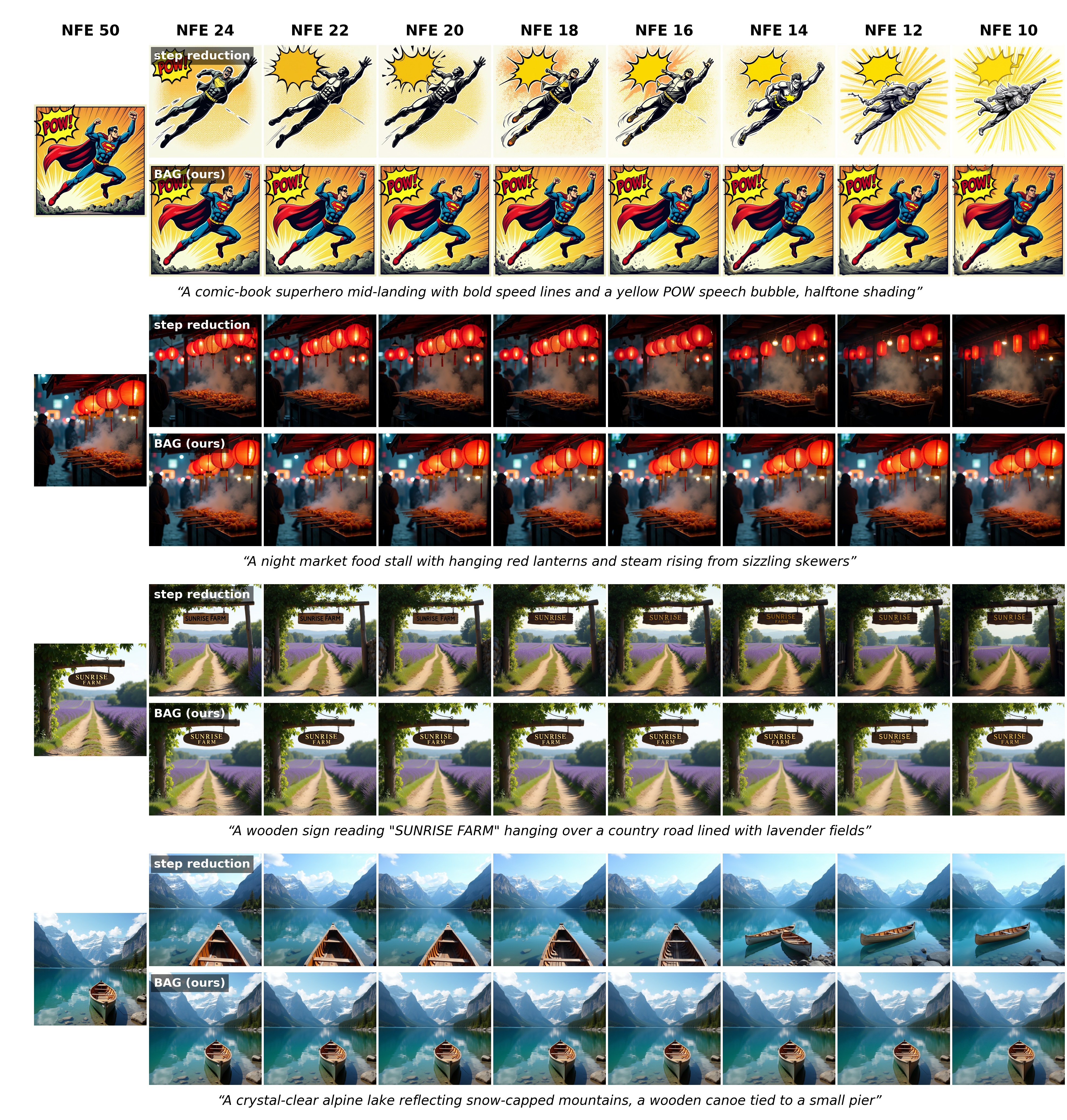}
\caption{\textbf{One \bag{} checkpoint across the budget range vs.\ naive step reduction} (four of the 50 sweep prompts; metrics over all 50 in \cref{tab:sweep}). Leftmost: the same-seed 50-step full-compute output. Columns run from NFE 24 down to 10; per case, the top row is naive step reduction at that step count and the bottom row is \bag{} at budget $B{=}$NFE on the 50-step grid. Step reduction loses text (the \textsc{pow} bubble, the second line of the \textsc{sunrise farm} sign), color, and object identity as steps shrink, and even at NFE 24 it settles on a different image, while the single \bag{} gate tracks the full-compute output across the whole range.}
\label{fig:sweep}
\end{figure*}

\begin{figure*}[t]
\centering
\includegraphics[width=\textwidth]{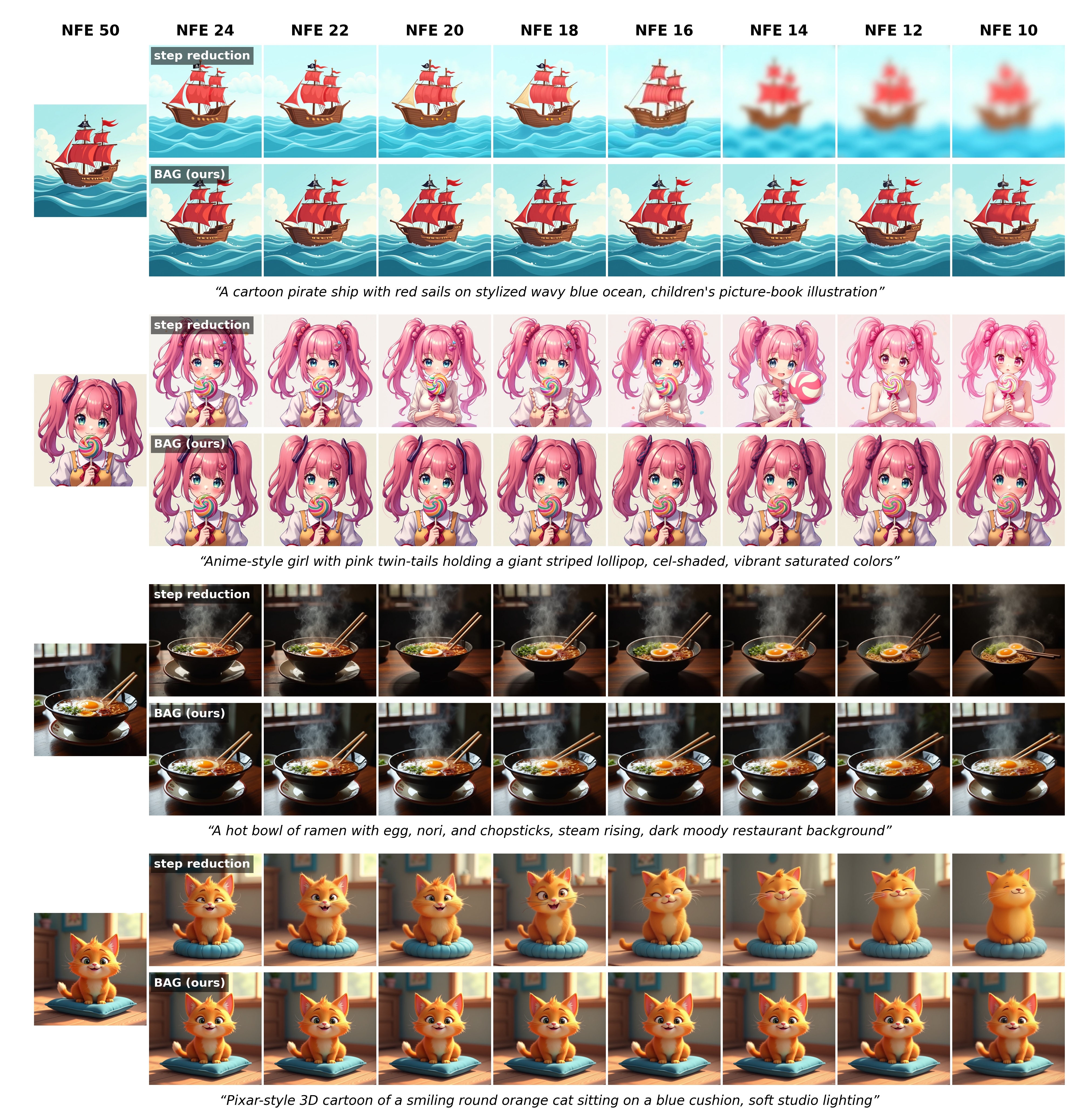}
\caption{\textbf{The same budget sweep, four further prompts} (layout and NFE grid as in \cref{fig:sweep}). As the budget shrinks, naive step reduction dissolves the pirate ship, drops the girl's lollipop, scrambles the ramen bowl's egg and steam, and softens the cat, while the single \bag{} checkpoint tracks the full-compute output down to NFE 10.}
\label{fig:sweep2}
\end{figure*}

\begin{figure*}[t]
\centering
\includegraphics[width=\textwidth]{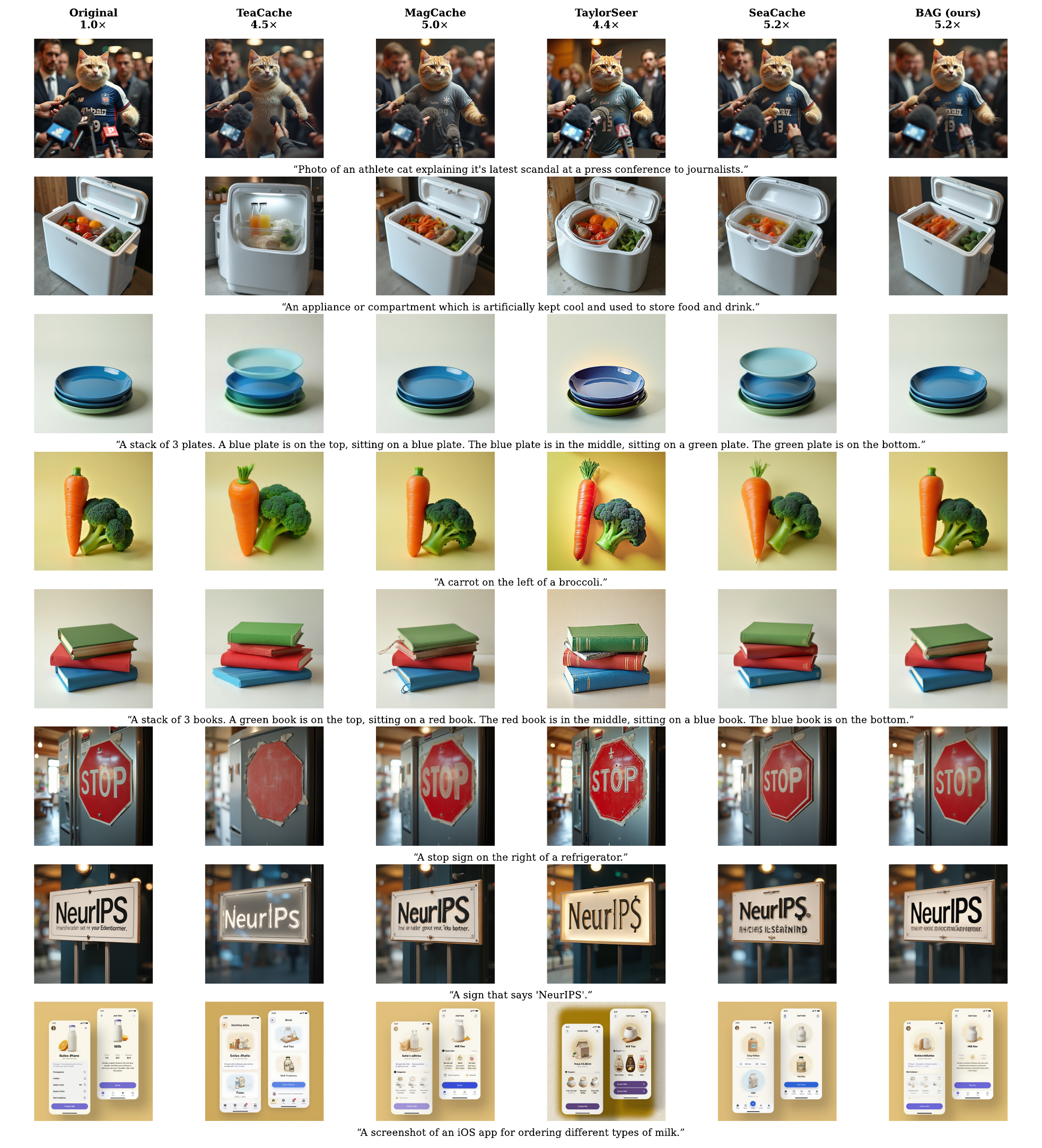}
\caption{\textbf{Additional qualitative results on FLUX.1-dev} ($\sim$5$\times$ tier, $B{=}9$; layout as \cref{fig:qualflux}). At matched NFE the heuristics erase sign text (STOP, NeurIPS), restyle objects, break layouts, and drift in object identity; \bag{} tracks the 50-step original.}
\label{fig:supp-flux}
\end{figure*}

\begin{figure*}[t]
\centering
\includegraphics[width=\textwidth]{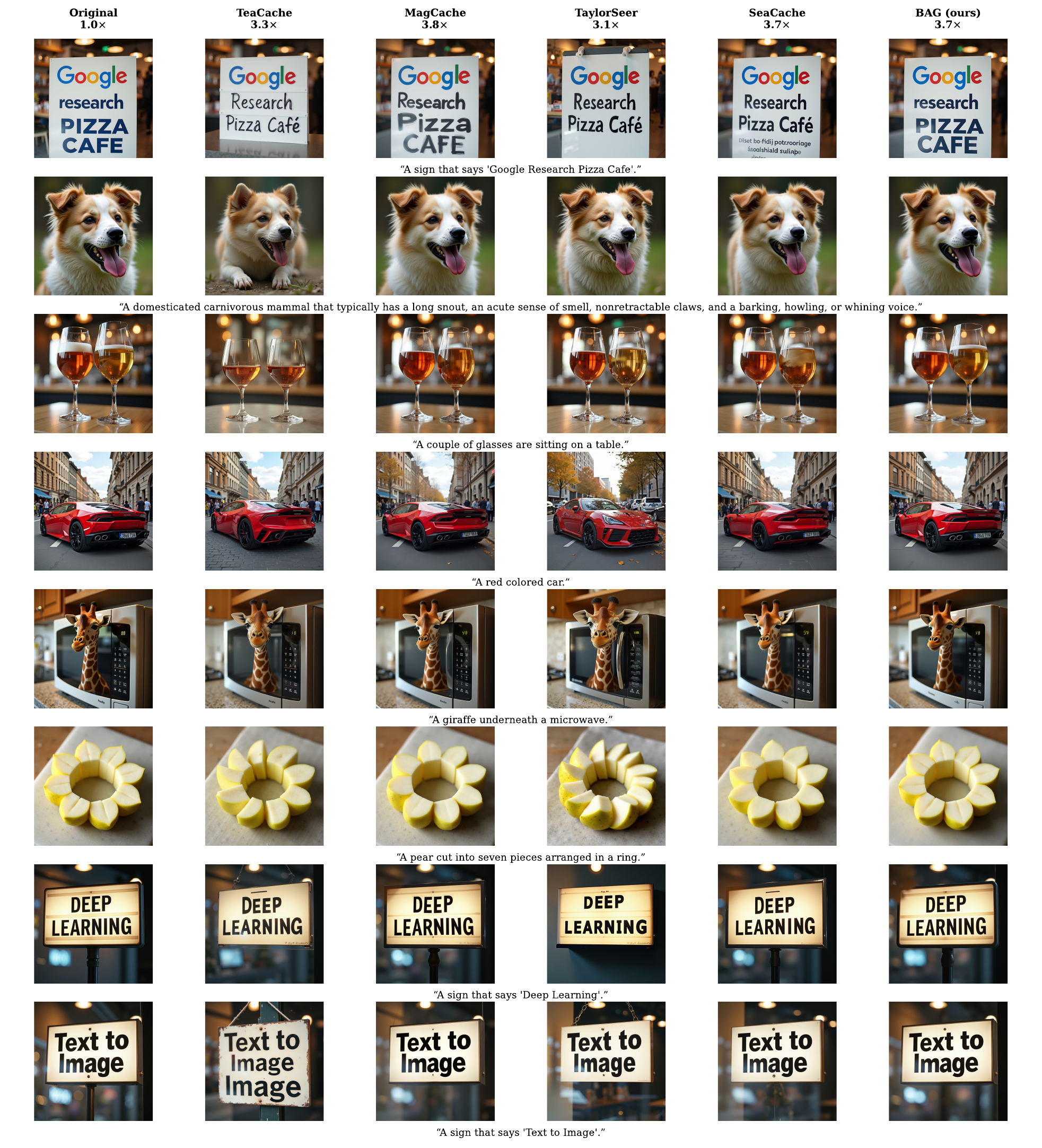}
\caption{\textbf{Additional qualitative results on FLUX.1-dev} ($\sim$3.8$\times$ tier, $B{=}13$; disjoint from the $\sim$5$\times$ set). At matched compute the baselines restyle sign text, change object pose and viewpoint, and deform layouts; \bag{} tracks the 50-step original.}
\label{fig:supp-flux13}
\end{figure*}

\begin{figure*}[t]
\centering
\includegraphics[width=0.86\textwidth]{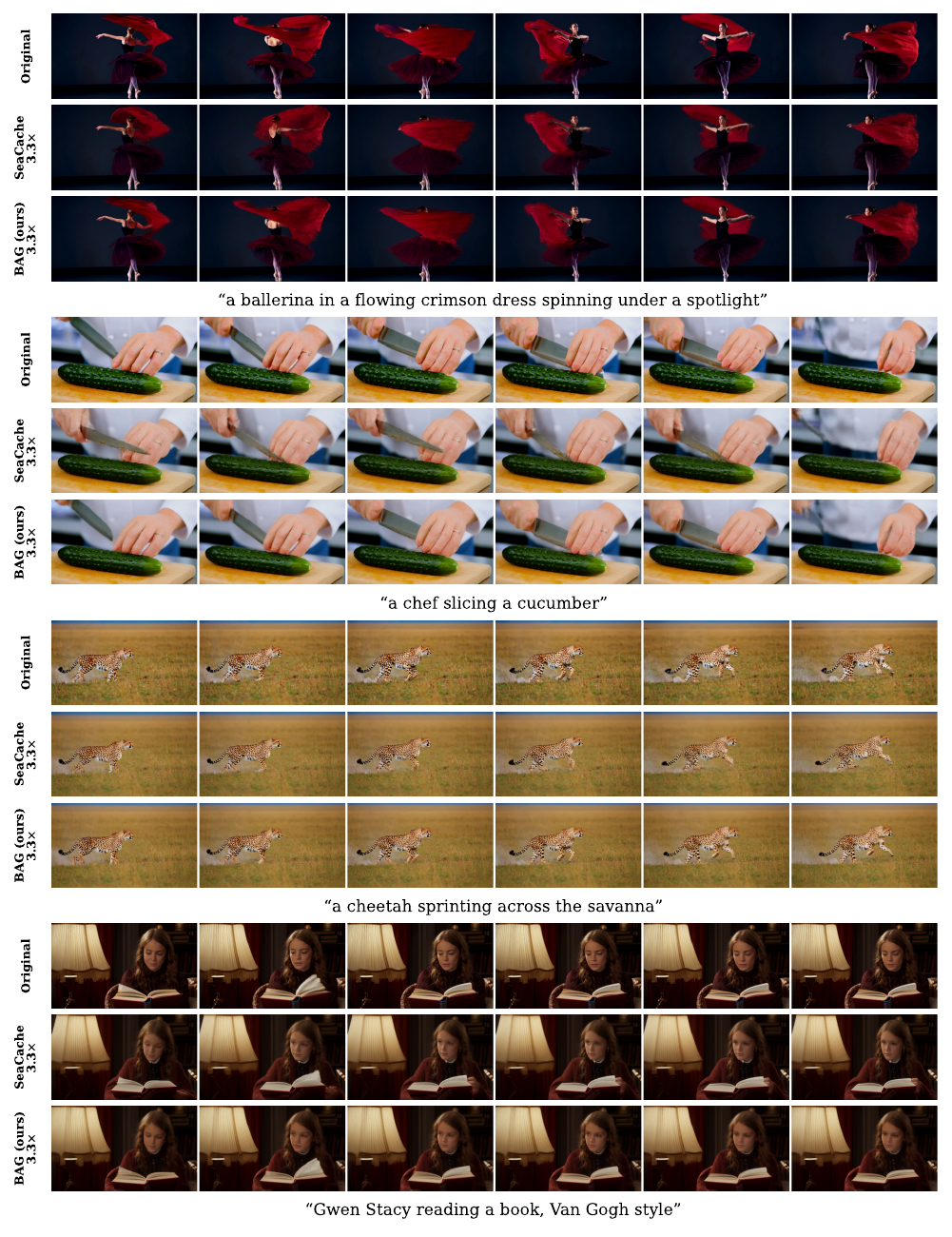}
\caption{\textbf{Additional qualitative results on Wan2.1} ($\sim$3.4$\times$ tier, $B{=}15$; six sampled frames). SeaCache drifts in framing and object identity; \bag{} tracks the 50-step original.}
\label{fig:supp-wan15}
\end{figure*}

\begin{figure*}[t]
\centering
\includegraphics[width=0.86\textwidth]{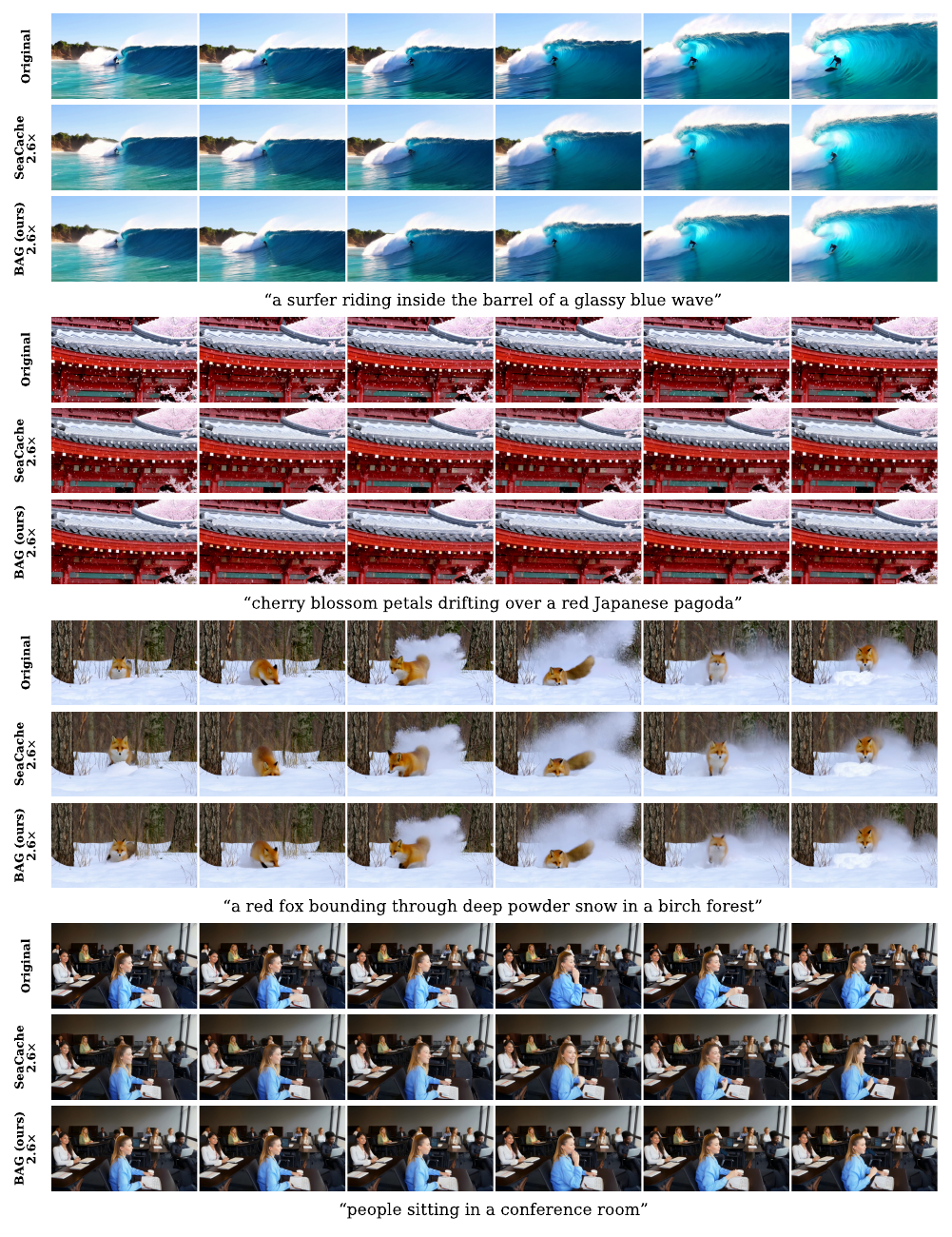}
\caption{\textbf{Additional qualitative results on Wan2.1} ($\sim$2.7$\times$ tier, $B{=}19$). The heuristic drifts in composition while \bag{} stays faithful.}
\label{fig:supp-wan19}
\end{figure*}

\cref{fig:masks} shows the realized compute placement of \bag{} and SeaCache at matched budgets. \cref{fig:sweep,fig:sweep2} compare one \bag{} checkpoint with naive step reduction over $B{=}10$--$24$. \crefrange{fig:supp-flux}{fig:supp-wan19} add prompts at two tiers on FLUX and Wan (FLUX at $B{=}9/13$; Wan at $B{=}15/19$). The pattern repeats across budgets and backbones: at matched NFE, threshold schedules lose object identity, geometry, and composition, while \bag{} stays closer to the full-compute output.

\section{Limitations}
\label{app:discussion}

Like other learned schedulers, \bag{} is not training-free: it needs an offline search to produce the reference schedules that the gate is trained on. This is the price of its budget control and per-prompt adaptivity, and it is paid once per backbone. The search scales with the number of label prompts, so it can be shortened at an acceptable quality cost.

\end{document}